\documentclass[lettersize,journal]{IEEEtran}

\usepackage{amsmath,amsfonts}
\usepackage{algorithmic}
\usepackage{array}
\usepackage{textcomp}
\usepackage{stfloats}
\usepackage{url}
\usepackage{verbatim}
\usepackage{graphicx}
\usepackage{balance}

\usepackage[english]{babel}

\usepackage{graphicx}
\usepackage{animate}
\usepackage{epsfig} 				
\usepackage{epstopdf}

\usepackage{listings}
\usepackage{color}
\usepackage{nameref}
\usepackage[hidelinks]{hyperref}
\usepackage[capitalize]{cleveref}
\usepackage{amsmath}	 			
\usepackage{amssymb}  				

\usepackage{dsfont}			
\usepackage{mathtools}

\usepackage{epigraph}
\usepackage{lscape}
\usepackage[]{nomencl}				
\usepackage{algorithm}
\usepackage{algorithmic}
\usepackage{multicol}
\usepackage{multirow}
\usepackage{etoolbox}

\usepackage{caption}
\usepackage{subcaption}
\usepackage{wrapfig}

\usepackage{siunitx}

\usepackage{floatflt}

\usepackage{url}

\usepackage{enumitem}

 \usepackage{cite}

\usepackage{bm}
\usepackage{booktabs}
\usepackage{tabularx}
\usepackage[table]{xcolor}
\usepackage{comment}
\usepackage{threeparttable}

\newcommand{\email}[1]{\href{mailto:#1}{\nolinkurl{#1}}}

\renewcommand{\sec}[1]{Section~\ref{#1}}
\newcommand{\fig}[1]{Fig.~\ref{#1}}

\newcommand{\app}[1]{Appendix~\ref{#1}}
\newcommand{\tab}[1]{Table~\ref{#1}}

\definecolor{lightgray}{gray}{0.9}

\newcommand{\titlelong}[0]{ Using Fiber Optic Bundles to Miniaturize\\ Vision-Based Tactile Sensors}

\newcommand{\sensor}[0]{DIGIT Pinki} 

\newcommand{\citet}[1]{\cite{#1}}
\newcommand{\citep}[1]{\cite{#1}}

\newcommand{\rev}[1]{\textcolor{black}{#1}}

\newcommand{\secondrev}[1]{\textcolor{black}{#1}}

\begin{document}

\title{\titlelong{}}

\author{Julia Di$^{1,2}$, Zdravko Dugonjic$^{3,4}$, Will Fu$^{1}$, Tingfan Wu$^{2}$, Romeo Mercado$^{2}$, Kevin Sawyer$^{2}$, \\Victoria Rose Most$^{2}$, Gregg Kammerer$^{2}$, Stefanie Speidel$^{4,5,6}$, Richard E. Fan$^{1}$, \\ Geoffrey Sonn$^{1}$, Mark R. Cutkosky$^{1}$, Mike Lambeta$^{2}$, and Roberto Calandra$^{2,3,6}$%

\thanks{$^{1}$ Stanford University, Stanford, CA, USA {\tt\small \{juliadi, jiaxiang, refan, gsonn, cutkosky\}@stanford.edu} \newline%
$^{2}$ Authors affiliated at time of work with Meta, Menlo Park, CA, USA \newline
        {\tt\small \{romeo12, kevin.sawyer, tingfan, victoriamost, greggk, lambetam\}@meta.com}%
        \newline%
$^{3}$ LASR Lab, Technische Universit\"at Dresden, Germany \newline {\tt\small \{zdugonjic, rcalandra\}@lasr.org}%
\newline%
$^{4}$ School of Embedded and Composite AI (SECAI)%
\newline%
$^{5}$ National Center for Tumor Diseases (NCT/UCC) Dresden \newline {\tt\small \{stefanie.speidel\}@nct-dresden.de}%
\newline%
$^{6}$ Centre for Tactile Internet with Human-in-the-Loop (CeTI)%
}
}

\markboth{IEEE Transactions On Robotics,~Vol.~X, No.~X, September~2024}%
{How to Use the IEEEtran \LaTeX \ Templates}

\maketitle


\begin{abstract}
	Vision-based tactile sensors have recently become popular due to their combination of low cost, very high spatial resolution, and ease of integration using widely available miniature cameras. The associated field of view and focal length, however, are difficult to package in a human-sized finger. In this paper we employ optical fiber bundles to achieve a form factor that, at 15\,mm diameter, is smaller than an average human fingertip. The electronics and camera are also located remotely, further reducing package size. The sensor achieves a spatial resolution of 0.22\,mm and a minimum force resolution 5\,mN for normal and shear contact forces. With these attributes, the \sensor{} sensor is suitable for applications such as robotic and teleoperated digital palpation. We demonstrate its utility for palpation of the prostate gland and show that it can achieve clinically relevant discrimination of prostate stiffness for phantom and \textit{ex vivo} tissue.
\end{abstract}



\section{Introduction}

\IEEEPARstart{F}{or} robot hands to substitute human hands in tasks requiring tactile acuity, they should match the force and spatial resolution of human fingertips while also having comparable stiffness and dimensions.
Towards this goal, we present a novel approach for miniaturizing vision-based tactile sensors by using fiber bundles as optical conduits, and demonstrate its use for manipulation tasks in constrained settings that cannot be done with larger fingers.

\begin{figure}
\centering
	\includegraphics[width=\columnwidth]{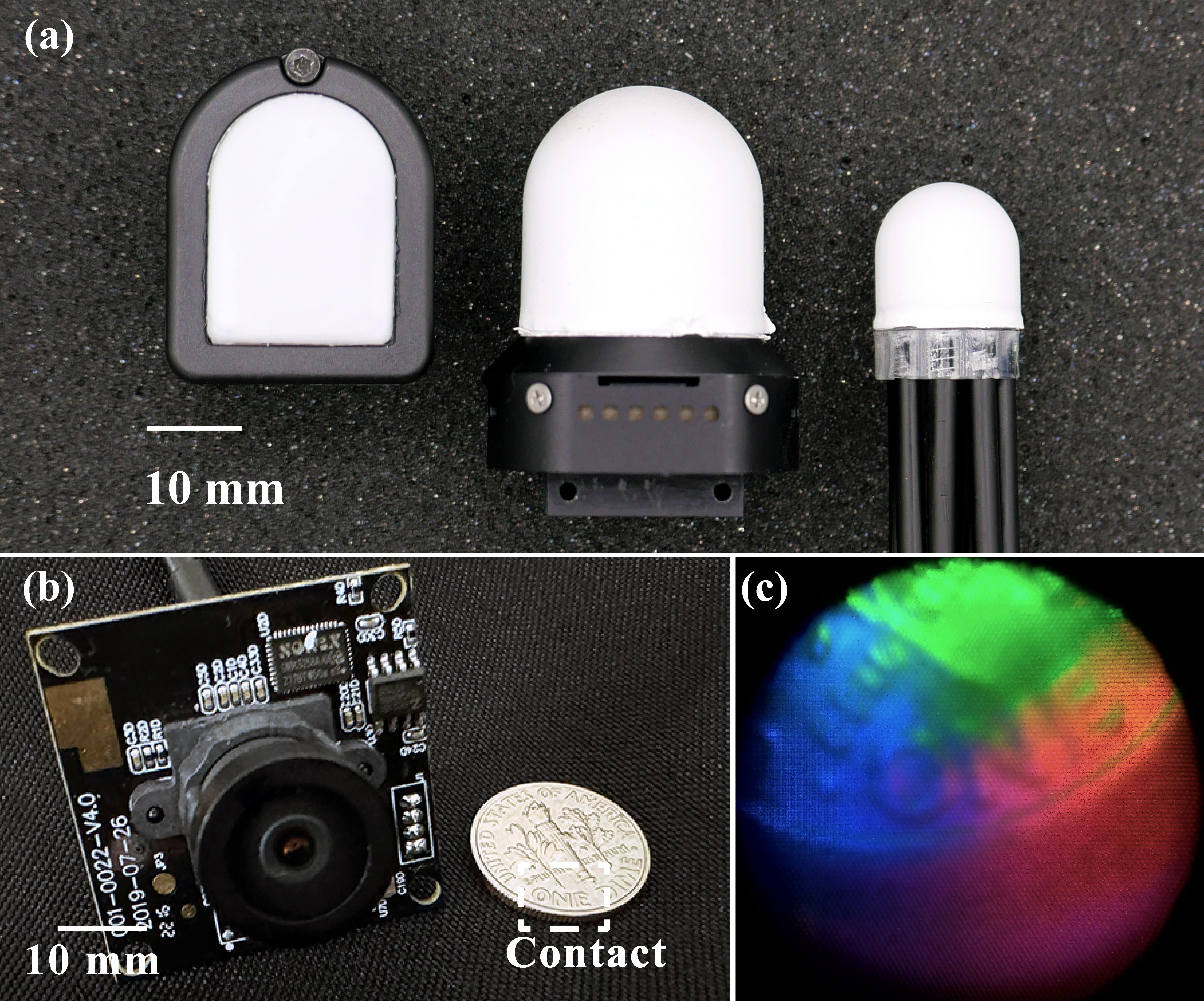}
	\caption{
    (a) DIGIT~\cite{lambeta2020digit} \textit{(left)} and DIGIT 360~\cite{lambeta2024digit360} \textit{(middle)} compared with the \sensor{} \textit{(right)} introduced in this work. The \sensor{}'s diameter of \SI{15}{\milli\meter} is achieved using optical fiber bundles as illumination and imaging conduits, thereby allowing the elastomer sensing element to achieve human scale. (b) \rev{\sensor{}'s off-the-shelf camera and supporting circuitry would be difficult to package in the base of a typical vision-based tactile sensor for the same tip size.} 
    (c) An image from \sensor{} pressed against the \rev{"ONE"} text \rev{of a U.S. dime}, with its sub-millimeter structure visible.}
	\label{fig:MainPhoto}
\end{figure}

Tactile spatial resolution matching or even exceeding that of the human fingertip has recently become possible using vision-based sensors. Popular examples include the DIGIT~\cite{lambeta2020digit}, GelSight~\cite{yuan2017gelsight}, and TacTip~\cite{lepora2021soft} sensors, which use miniature cameras to capture the deformation of a transparent elastomer gel and apply machine learning techniques to estimate the associated surface deformation and stresses. These examples, however, are relatively large compared to human fingertips, chiefly due to the need to meet the optical path requirements of an internal camera.
Several researchers have recently proposed slimmer designs to address this issue, but
it remains a challenge to produce designs that match the geometry of a human fingertip.
For example, the GelSlim~3.0~\cite{taylor2022gelslim} and GelSight Wedge~\cite{wang2021gelsight}, while slim, lack a rounded three-dimensional sensing surface. Meanwhile, dome-shaped, three-dimensional sensors such as OmniTact~\citep{Padmanabha2020OmniTact}, GelSight360~\cite{tippur2023gelsight360}, and Insight~\cite{sun2022soft} have base diameters of \SI{30}{\milli\meter}, \SI{28}{\milli\meter}, and \SI{40}{\milli\meter} respectively. At the size of a large adult thumb or toe, they are not a direct substitute for a human fingertip. Moreover, it is difficult to miniaturize them further due to the focal length and field of view of the camera, and the desire to package the camera and all associated electronics, including illumination, in the base.

Slim and sensitive fingertips in the size range of the smallest adult human fingers unlock the kinds of manipulation tasks that cannot be done with larger fingers. One potentially life-saving example is cancer detection, through the tissue palpation of internal cavities during remote or teleoperated exams. In particular, prostate cancer is the fifth leading cause of cancer death in men worldwide but can be detected early with regular screening protocols including the digital rectal examination~(DRE), in which the prostate area is gently palpated with a gloved finger through the rectum~\cite{wang2022prostate}. Although the exam is easy to perform, the result is subjective and depends on the experience of the examiner~\cite{smith1995interexaminer}. A robotic care provider, perhaps controlled remotely, could provide objective and quantitative measurements if equipped with a slim and sensitive tactile fingertip.

Motivated by the problem of palpation in internal cavities and similarly constrained environments, we present a design approach that can miniaturize vision-based tactile sensors to match average human fingertips, shown in \fig{fig:MainPhoto}. At \SI{15}{\milli\meter}, the diameter of the \sensor{} is no larger than an average female index fingertip or a 5th percentile male fingertip \cite{garrett1971adult}. 
Inspired by early fiber-based imaging research, we use optical fiber bundles as conduits for illumination and imaging, allowing the separation of the sensing element from supporting circuitry. 

Because the supporting circuitry and processing are located remotely, the sensor size is no longer constrained by the base and the optical path for the internal camera, lens, and lighting. Instead, the primary constraint is the bundle size, as determined by the desired resolution and lighting requirements. 

In presenting this new design, our contributions are:
\begin{itemize}[noitemsep,topsep=0pt]

    \item the design, fabrication, and testing of an optical tactile sensor that uses optical fiber bundles to match human fingertip dimensions and approach human tactile sensitivity;
    \item an optical system design that integrates fibers \rev{as conduits} for imaging and illumination \rev{to separate the sensing element from supporting circuitry};
    \item a demonstration of the suitability of the sensor for medical palpation and an ability to provide clinically relevant levels of tissue stiffness discrimination.

\end{itemize}
To encourage further experimentation with the presented technology we are also open-sourcing the hardware design at \url{https://github.com/facebookresearch/digit-design}.


\section{Related Work}
\label{sec:related}

\subsection{Vision-Based Tactile Sensors}
In recent years, camera technology has improved significantly, becoming smaller, cheaper, and higher in resolution. Such improvements have opened up new implementations of tactile sensors that cast tactile sensing as a computer vision problem---learning or modeling tactile features or properties from high-resolution images captured by a camera. When combined with significant progress in computer vision and machine learning, these new high-resolution vision-based sensors are promising for tactile perception.

A prevailing design paradigm for vision-based tactile sensors involves the use of an optically-transparent elastomer, a camera, and a lighting system. These types of sensors then image the indentation an object makes on the elastomer at the point of contact, thereby capturing surface property information as a tactile image. The information can include the texture, contact area, forces, contact centroid, and other computed tactile features that may then be used for manipulation.

To better suit dexterous manipulation tasks, recent works have developed these sensing designs from planar to a three-dimensional. One common form factor that has emerged is an elastomer gel shaped in a hemisphere or dome, reminiscent of a human fingertip that may then be mounted on a robotic finger. OmniTact~\cite{Padmanabha2020OmniTact}, GelTip~\cite{gomes2022geltip}, and GelSight360~\cite{tippur2023gelsight360} are all recent examples of these fingertip vision-based tactile sensors \rev{with an internal rigid shell that provides additional structural support. A hemispherical shape may also be achieved without an internal rigid shell, of which Soft-Bubble~\cite{alspach2019soft} and DenseTact~\cite{do2022densetact} are examples}. Another common theme is that of a slim elastomer gel for direct integration with a gripper, with elastomers shaped as wedges~\cite{taylor2022gelslim}, finrays~\cite{liu2022gelsight}, or fingers themselves~\cite{sun2022soft,zhao2023gelsight}.

Despite the differences in elastomer shapes,  all of these sensors either house the camera internally at the base of the sensor, or use an arrangement of mirrors or 

prisms which direct light to an internal camera located a short distance away. Further miniaturization of the sensing element may prove challenging with these existing designs, which leads us to investigate a different design solution using optical fiber bundles. 

\subsection{Fiber-Based Tactile Sensing}
Fiber-based sensors have a long history in applications requiring some or all of the following considerations: small size, light weight, immunity to electromagnetic radiation, or an ability to transfer information over long distances \cite{zeng2014fiber,taffoni2013optical}. In particular, optical fiber bundles are widely used for light delivery and, in the case of coherent fiber bundles (CFBs), direct image delivery as well. Both coherent and incoherent fiber bundles have been used in borescopes and endoscopes for decades, allowing access to the smallest vessels of the human body \cite{lee2010scanning,orth2019optical}. These characteristics, and the ready availability of high-quality coherent fiber bundles, recommend them for lighting and image delivery in our application.

We note also that the general idea of imaging through fiber-based optical waveguides has already been explored in early vision-based tactile sensors, in part because camera miniaturization was not yet mature~\cite{shah2021design,heo2008tactile,maekawa1993finger}. 
A seminal example from 1988 involved a finger-shaped optical tactile sensor where an optical fiber waveguide conveyed a tactile image to a charge-coupled device (CCD) camera~\cite{begej1988planar}. In this design, each fiber is attached to a transducer membrane along a hemispherical surface, and the tactile image is created when an object scatters light at the contact point due to frustrated internal reflection (FTIR). A drawback to the FTIR approach is that some of the details of the contact image are lost in order to enhance the contrast between contacting and non-contacting areas.
More recent sensors with fiber-based conduits have also exploited FTIR but pose challenges for manufacturing at scale (e.g. \cite{xie2013magnetic}) or provide a substantially lower resolution (e.g., \cite{ali2012characteristics, yussof2010sensorization}) than what is available from modern cameras.

In this work, we revisit the idea of fiber-based tactile sensing through the lens of endoscopic design principles to achieve high spatial resolution with small sensor size. Modern incarnations of flexible fiberoptic endoscopes often employ a proximal video camera for image capture through the CFB~\cite{lee2010scanning}, or use a miniature CCD chip at the distal end of a flexible shaft, with incoherent optical fiber bundles to deliver diffuse white-light illumination from the proximal to distal end~\cite{zouridakis2003biomedical}. Both designs result in a slim imaging scope, with trade-offs between size and resolution. While previous research in fiber-based imaging has investigated optical techniques (e.g., imaging relay and scanning mechanisms~\cite{boppart1999optical},  fiber characterization \cite{wood2017fiber}, and case studies in medicine \cite{rouse2004design,elahi2011future}), we do not know of works that extend these design details to the system level needed for integration into vision-based tactile sensors. Such details include the arrangement of fibers and lighting for a hemispherical elastomer dome and the design of a miniature distal lens for imaging at close range with high field of view.

\subsection{Tactile Applications in Medicine}
Tactile sensing has been explored for a number of applications in medicine. In the realm of minimally invasive surgeries, tactile feedback systems enhance precision by providing surgeons with a sense of touch~\cite{tiwana2012review,Othman2022Tactile, huang2020tactile}. Other researchers have used tactile feedback systems to improve robotic manipulation tasks, such as automatic swab sampling~\cite{wang2020design,Li2023Visuotactile}. Tactile sensors also are useful as diagnostic tools, for instance in the classification of tissue mechanical properties for patient examination~\cite{konstantinova2014implementation,Jia2013Lump,laufer2015sensor}. 

In this work we demonstrate the benefits of \sensor{}'s small size in a medical palpation case study of the prostate gland. Over a million men receive a prostate cancer diagnosis each year, with many physicians routinely screening via the digital rectal exam (DRE) ~\cite{naji2018digital}. In the DRE, a physician inserts a gloved finger to probe the lower rectum, dynamically palpating for any abnormalities in the prostate gland. Because around 80\% of prostate cancers arise in the posterier region~\cite{bott2002anterior}, which is palpable via the readily-available and inexpensive DRE, it continues to be recommended for clinical practice in addition to more advanced biomarker, imaging, or biopsy diagnostic tools~\cite{wei2023early}.

Nevertheless, because the DRE and similar palpation examinations are widely accepted as \emph{subjective} measurements, researchers have investigated ways to quantify them with sensorized robotic systems. To mimic the DRE, some researchers have affixed a probe to a force-torque sensor for palpation case studies, or have used fibers with Fiber Bragg Gratings (FBGs) for strain sensing~\cite{ahn2014robotic, tanaka2000development,iele2021miniaturized}. \rev{Although these efforts are promising, clinicians today practice a variety of palpation techniques involving both normal and shear forces to obtain a wealth of information about the tissue (e.g. firmness, texture, temperature), which a simple 1-DOF force sensor may not be able to capture.}
Other researchers have investigated robotic palpation classifiers for breast and prostate tumors~\cite{sanni2022deep,nichols2015methods}. None of these investigations has demonstrated a high-resolution vision-based tactile sensor small enough to be clinically relevant for the DRE and similarly constrained manipulation tasks.
	

\section{Design and Fabrication} 
\label{sec:design}

	\begin{figure*}[ht]
\centering
	\includegraphics[width=1.0\linewidth]{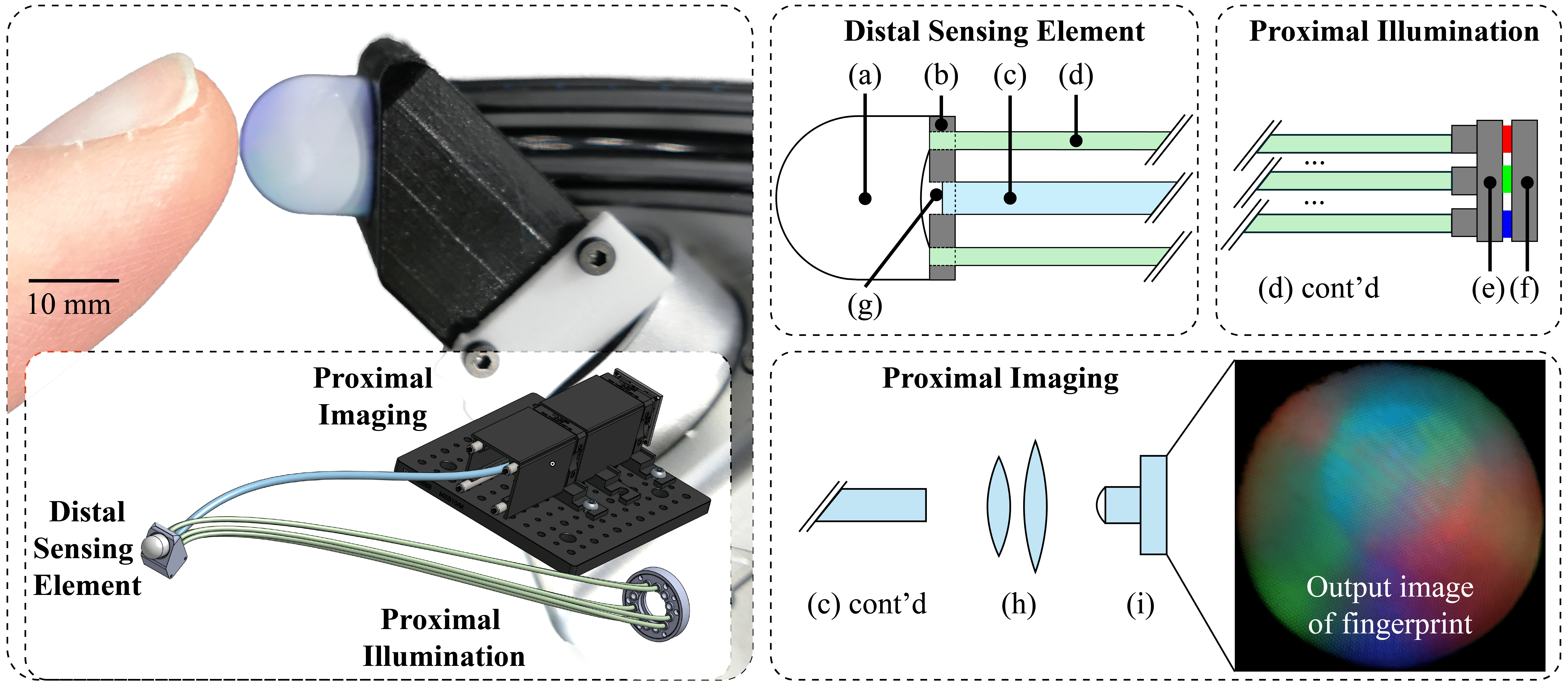}
	\caption{The proposed system consists of: a distal sensing element, a proximal imaging system, and a proximal illumination system, all of which are connected with optical fiber bundles. The distal end contains (a) an elastomer gel mounted in (b) a 3D-printed housing (pictured at left is a housing compatible with the Allegro Hand, but this housing could be designed for other uses). Both the (c) imaging and (d) illumination fiber bundles mate to the gel. At the proximal end, the illumination fiber bundles mate to the (e) collimator coupled to (f) LEDs. When making contact with an object (e.g. a woman's index finger), the gel is first imaged with (g) a distal lens, conveyed through the imaging fiber bundle, then magnified by (h) optical lenses and captured by the (i) camera. The proximal imaging and illumination systems may be co-located or separate.
    }
	\label{fig:sensorDiagram}
\end{figure*}

This section details the design principles and corresponding choices made during the design of the \sensor{}.
Descriptions of how each of the components of the sensor works and their manufacturing process are presented. 

\subsection{System Design}
\label{subsec:system}

We identified three primary requirements to guide the design:
\begin{itemize}[noitemsep,topsep=0pt]
    \item small elastomer sensing element with diameter and stiffness comparable to a human fingertip,
    \item high spatial resolution for computer vision,
    \item high field of view across the curved fingertip area.
\end{itemize}
Based on these requirements, we explored design details for the elastomer gel and interface, the illumination via fibers, and the imaging via fibers. For the proof-of-concept prototype, we constructed a prototype that is \SI{15}{\milli\metre} in diameter, which is shown in \fig{fig:sensorDiagram}. As noted earlier this makes \sensor{} approximately the size of an adult female index fingertip or a fifth percentile male fingertip~\cite{garrett1971adult}. 
Furthermore, as discussed in \cref{sec:discussion and future work}, the design can in principle be reduced to just a few millimeters in diameter.

\subsection{Elastomer Gel}
\label{subsec:elastomer}

\sensor{} uses a deformable elastomeric material in the shape of a hemispherical dome atop a short cylindrical section. A detailed manufacturing guide and open-source resources can be found on \url{https://github.com/facebookresearch/digit-design}. 

A hemispherical gel acts as an integrating sphere for light rays, so a fully Lambertian scattering surface reduces indentation contrast compared to a Gaussian scattering surface~\cite{lambeta2024digit360}. Therefore, to control the light scattering at the surface, we machined a metal mold with a smooth and polished surface and then finely sandblasted the surface to achieve a precise texturing. We also found alternatively that one may cast a dome-shaped daughter mold with Mold Star or a similar mold-making silicone and vary the amount of mold release used on the mold in order to achieve a similar texturing, though with less consistency than the machined mold. 
Once a mold was prepared for casting, Smooth-On Solaris (Shore 10A), an optically clear silicone, was syringed into the mold cavity. We vacuum degassed the silicone while in the mold to ensure that no air bubbles were present. All samples are cured for 24 hours at room temperature. 

There are two layers on the elastomer gel dome: a reflective layer and a protective layer over the reflective layer. After demolding, the gel dome was cleaned with 99\% isopropyl alcohol, and airsprayed with a thin layer of Inhibit-X and then left to outgas for 30 minutes at room temperature. Then we sprayed Rustoleum Mirror Effect paint until the surface is evenly coated with a reflective layer. This was followed by a thin layer of Inhibit-X to prepare the surface for the protective layer. 
To make the protective layer, we mixed Ecoflex 00-10 in a ratio of 1:1 Part A to Part B, with an additional 1\% white Smooth-on Silc Pig by weight. The mixture was then diluted with 4\% by weight of NOVOCS matte. The mixture is vaccuum degased and then drizzled onto the reflective layer in a process akin to drip candy coating. This drip process took only a few minutes and was much faster than airspraying silicone, a fabrication technique in other work that can take hours, or painting with a small brush. The resulting thin layer of protective Ecoflex 00-10 silicone was fully cured after 4 hours. At this point, the elastomer gel dome was ready to use.

\subsection{Illumination}

With the proposed fiber-based approach, \sensor{} used plastic optical fibers coupled to LEDs for illumination rather than LEDs directly. The light source we used in this prototype is an Adafruit Neopixel ring with 5050 RGB LEDs, which is a Lambertian source according to the manufacturer radiation diagram. We then directly coupled each LED package to a 48-core incoherent plastic optical fiber bundle, with end faces that were manually polished using 200, 400, and 1000 grit sandpaper. The bundles were directly coupled to the elastomer gel dome and held with a 3D-printed housing (\cref{fig:sensorDiagram}). 

Consideration was given to the location of the illumination fibers relative to the base of the elastomer gel dome. We defined their placement to the elastomer gel with these parameters: number of fiber bundles $n_b$, radius of each fiber bundle (inclusive of any jackets) $r_f$, the radial distance between the center of each bundle from sensor center $r_b$, and angular spacing between bundles $\alpha_b$. While 3 was a minimal number of bundles for RGB illumination \rev{(the color channels provide shadow and contrast to aid models to learn from an image), the following parameters empirically worked to provide illumination coverage across the field of view: $n_b = 6$, $r_f = \SI{3}{\milli\meter}$, $r_b = \SI{5.75}{\milli\meter}$, and $\alpha_b = 60^{\circ}$}.

Some important design parameters when choosing a fiber-based illumination conduit are the core diameter $c_d$ and the numerical aperture of the fiber $\mathit{NA}$. These parameters help characterize both how much light is relayed through the bundle and the output light profile from the bundle. 

We wanted to minimize the light loss through the fiber bundles. We chose a \SI{3}{\milli\metre} diameter incoherent fiber bundle to minimize geometry-based coupling loss with the LEDs while maintaining a small overall diameter at the sensing element. We also used an index-matched gel when optically coupling to minimize Fresnel loss. Therefore, any light loss in our system is largely angular coupling loss, which can be estimated. Based on Lambert's cosine emission law, $I(\theta) = I_{0}\cos \theta$, we can calculate the angular coupling loss as
\begin{equation}
\eta_{\mathit{ang}} = \sin \theta_{\mathit{fiber}}^{2}\,,
\end{equation} 
where the fiber acceptance angle is $\theta_{\mathit{fiber}} = \arcsin(\mathit{NA})$, For the proof of concept prototype in this work, we calculated a 25\% loss due to angular acceptance coupling loss. For any future prototypes that use much longer fibers, the loss calculation should include attenuation loss. The fiber used in \sensor{} has a 0.65 \unit{\decibel}/\unit{\metre} attenuation rating, but because the fibers used in this prototype are only about \SI{0.4}{\meter} in length, we ignored attenuation loss for this case. Fiber system light loss may also be partially compensated for by varying the intensity of the input LEDs, \rev{though care must be taken to avoid providing too much background illumination which reduces the sensitivity of the sensor}.

\begin{figure}[t]
\centering
\subfloat[Illumination without fiber]{\includegraphics[width=0.5\columnwidth]{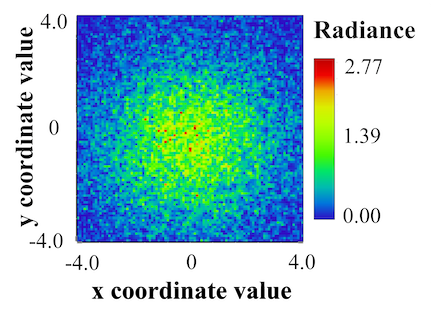}%
\label{fig:illuminationWithoutFiber}}
\hfil
\centering
\subfloat[Illumination with fiber]{\includegraphics[width=0.5\columnwidth]{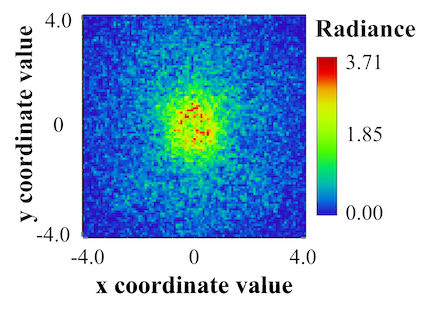}%
\label{fig:illuminationWithFiber}}
\hfil
	\caption{Comparison of the ray-traced outputs in Zemax Optics Studio of lighting from a 5050 RGB LED package (\subref{fig:illuminationWithoutFiber}) without and (\subref{fig:illuminationWithFiber}) with the fiber bundle. The fiber output is narrower and more focused.}
	\label{fig:zemaxPlot}

\end{figure}

One result from using fiber bundles for lighting is that the output light profile is different from a Lambertian diffuse source. To study this, we modeled in Zemax Optics Studio a comparison between the 5050 LED package without fibers and with the \SI{3}{\milli\metre} diameter PMMA fiber bundle as a lighting conduit~(\fig{fig:zemaxPlot}). In this model, we plotted the ray-traced light intensity received at a rectangular detector \SI{3}{\milli\metre} away from the light source using 1,000 analysis rays for both scenarios. We found that illumination through a fiber bundle is more narrow in output compared to the Lambertian light source of a direct LED. We then approximated the lighting as a Gaussian beam instead of a Lambertian source, with a cone of light given by the numerical aperture ($\mathit{NA}$) of the fiber bundle. 

The amount of Gaussian scatter affects the image contrast and therefore sensitivity of the sensor. Very low amounts of Gaussian scatter may result in non-uniform background illumination and specular glints on the sides of the sensor that could saturate the vision system, which is not necessarily disadvantageous. However, to achieve a high contrast across the field of view of the sensor, some degree of Gaussian scatter is needed for imaging contrast across the field of view. Accurate scattering can be realized with proper tool machining; sandblasting the mold is a more accessible method with adequate results.

\subsection{Imaging}

We chose the IMX298 16MP Arducam camera with a pixel size of 
1.12\,×\,\SI{1.12}{\micro\metre} and a maximum image resolution of 4656\,×\,3496 pixels when imaging at \SI{10}{\hertz}, though the frame rate is adjustable up to \SI{30}{\hertz} at lower resolutions. The camera parameters such as brightness, white balance, and exposure are also adjustable. \rev{At 38\,x\, \SI{38}{\milli\meter}, this particular camera would be too bulky to be deployed at the base of the gel dome in the style of the more typical vision-based tactile sensors.}

For imaging through a fiber bundle, the effective spatial resolution is most often constrained by the fiber packing density and fiber core diameter, rather than the camera sensor and imaging optics. Resolution in this case refers to the minimum feature dimensions that can be resolved with a modulation transfer function (MTF) $\geq$ 0.5. The proof of concept used a 7400-core imaging fiber bundle made of Poly (methyl methacrylate) plastic, which was packed as a hexagonal array in a \SI{2}{\milli\metre} diameter polyethylene jacket and had a individual fiber core of \SI{0.2}{\milli\metre}.

To achieve a high field of view of the hemispherical gel, a wide-angle lens is necessary at the distal end of the fiber. Most existing vision-based tactile sensors use a commercial off-the-shelf camera module with a M12 wide angle lens, but these lenses are too bulky for our purposes. 

For tests described in the next section, we only image the tip (\SI{60}{\degree} field of view) with a non-hyperfisheye distal lens. However, it is possible to substitute in a hyperfisheye lens, assuming it could be obtained at moderate cost, which we leave for future iterations of this work.

\subsection{Assembly}

The prototype, as pictured in \fig{fig:sensorDiagram}, was assembled in the following manner. 

First, we prepared the distal end of the sensor. We bonded a completed gel to a thin 3D-printed thread using Smooth-On Sil-Poxy Silicone Adhesive. Once cured, we then were able to screw the gel onto a corresponding 3D-printed gel housing. A few gel housing configurations were designed during prototyping, but for all reported experiments in the next section we used a 3D-printed housing compatible with the fingertip of an Allegro hand. Other vision-based tactile sensors typically overmold the gel elastomers directly onto a base PCB or skeleton frame, but we found that a ``screw-on" gel allowed for easier swapping of gel tips, which is especially useful for medical applications. Of note is that this modularity is in part enabled by the fiber design, where the camera and other sensitive opto-electronics are packaged remotely, so swapping a gel only involves unscrewing the old gel, reapplying an index-matched optical gel, and screwing a new gel back on.

In the prototype, we used a MEDIT, Inc. fiber bundle containing 7,400 cores and a \SI{60}{\degree} distal lens for the imaging conduit, and a 48-core incoherent optical fiber bundle for the illumination conduit. These illumination and imaging fiber bundles were directly coupled to the gel with an index-matched optical gel (Cargille 0608), and adhered with cyanoacrylate to the gel housing from the opposite side. 

Next, we prepared the proximal end of the sensor, consisting of the proximal illumination and imaging systems. The illumination and imaging systems may be co-located, but in the prototype built in this work they were separated \rev{for easier access to the individual components for debugging}. 

For the illumination system, the illumination fibers were adhered with cyanoacrylate to a 3D-printed collimator, which was directly coupled to a 12-LED Adafruit Neopixel ring mounted in an LED holder. The illumination fibers have minimum bend radius of \SI{75}{\milli\metre} before light leakage occurs, so we ensured during operation that the fibers stay within the acceptable bend radius to prevent further light loss. 

To image the proximal end face of the imaging fiber bundle, we set up an optical bench system with ThorLabs components. All pieces were mounted to an optical breadboard, which we set up with a \SI{30}{\milli\meter} adjustable optical cage and light shields. From the imaging fiber bundle, we aligned an adjustable diopter, a 10x Plan microscope objective, and a microscope eyepiece to a 16MP \rev{Arducam} IMX298 USB camera connected to a computer. \rev{The lens of the Arducam IMX298 USB camera was focused manually before assembly. The adjustable diopter was used for fine focusing of the output image, and the placement of the microscope objective and eyepiece on the optical breadboard was also adjusted to focus the output image.}

\app{app:CAD} provides the CAD for this assembly, \app{app:assembly} provides more discussion about the assembly process, and \app{app:manufacturingGuide} provides the materials and components.


 \section{Sensor Characterization} 
\label{sec:characterization}

	This section describes the characterization results for the \sensor{} in terms of spatial resolution and normal and shear force estimation. 
We also provide an analytic validation of the contact patch evolution using contact mechanics theory.

\subsection{Spatial Resolution}

\begin{figure}[t]
\centering
    \includegraphics[width=\columnwidth]{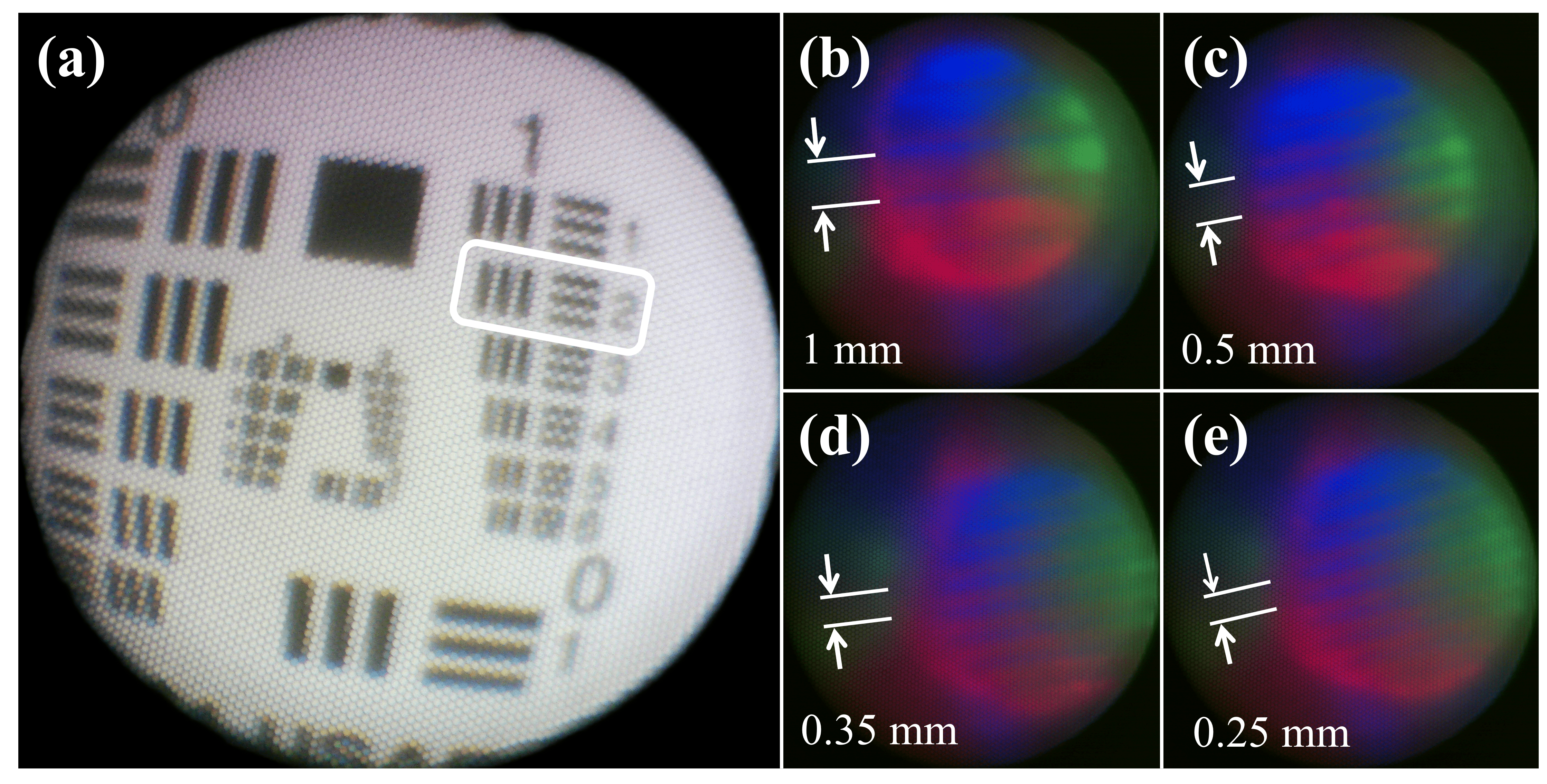}
\caption{(a) The USAF 1951 Test Pattern resolution target as captured with the imaging fiber (no gel). The working distance is set to the height of the gel. Group 1, element 2, which is outlined in white, can still be resolved (lines do not blur). When zoomed in to the output image, there is an apparent "honeycomb" pattern, as is typical when imaging through fibers. (b-e) Readings from a \sensor{} when pressed against machined calibration blocks with known line widths of: (b) \SI{1}{\milli\meter}, (c) \SI{0.50}{\milli\meter}, (d) \SI{0.35}{\milli\meter}, and (e) \SI{0.25}{\milli\meter}. The indentations of the lines are visible.}
\label{fig:resolution}
\end{figure}

\begin{table}[] 
\begin{threeparttable}
    \small
	\centering
	\renewcommand{\arraystretch}{0.8}
	\caption{Comparison of DIGIT, DIGIT 360, and \sensor{}}
     \label{tab:DigitComparison}
 
\begin{tabularx}{\columnwidth}{ l | *{5}X } 
 \toprule
        {\textbf{Sensor}} 
        & \textbf{Sensing Area} (\unit{\milli\metre^{2}}) & \textbf{Sample Rate} (\unit{\hertz}) & \textbf{Spatial Resolution} (\unit{\milli \metre}) & \textbf{Normal Force Resolution} (\unit{\newton}) & \textbf{Shear Force Resolution} (\unit{\newton})
    \\
    \midrule
    
  DIGIT & 304 & 60 & 0.150 & 0.006 & 0.012 
    \\
  DIGIT~360 & 2,340 & 240 & 0.007 & 0.0010 & 0.0013 
   \\
  \sensor{} & 181\tnote{*} & 10\tnote{*} & 0.22 & 0.005 & 0.005
   \\
   \bottomrule
\end{tabularx}

\begin{tablenotes}\footnotesize
\item[*] Calculation with $60^\circ$ distal lens.
\item[**] Sampling rate at highest resolution of 4656×3496 pixels. A \SI{30}{\hertz} sampling rate is available at lower resolutions (1920×1080 pixels)\rev{, which is the frequency and resolution used for data collection in this work}.

\end{tablenotes}
\end{threeparttable}
\end{table}

To verify the resolution of the sensor system optics, we used a standard United States Air Force (USAF) resolution target as commonly used for microscopes and cameras. This benchtop test is a tri-bar test, where the optical resolution of the system is determined by assessing the visibility of groups of bars. The resolution is the smallest group where all three black bars can be separated. The image from the USAF resolution target test captured through the fiber bundle is shown in \fig{fig:resolution}a with a working distance equal to that at the tip of the gel. We clearly observed Group Number 1, 
Element Number 2, for a resolution of 2.24 lines per \SI{}{\milli\meter}. Based on this resolution test target, the \sensor{} can resolve a line width of approximately \SI{222.72}{\micro\metre}. The true resolution limit is between \SI{222.72}{\micro\meter} and \SI{198.43}{\micro\meter} (which corresponds to Element Number 3). Note that this resolution limit is for a system using the \SI{60}{\degree} distal lens and 7,400-core imaging fiber; changes to either optical component could improve the resolution limit.

We also tested the resolution of imaging through an elastomer gel. To do so, we custom-machined a calibration test target block. The block has line widths of \SI{1.00}{\milli\meter}, \SI{0.5}{\milli\meter}, \SI{0.35}{\milli\meter}, and \SI{0.25}{\milli\meter}. The results from this test are shown in \fig{fig:resolution}b-e. The camera exposure in this sequence is set to $-1$. The indentations of the lines are clearly visible down to \SI{0.25}{\milli\meter}. 

\subsection{Normal and Shear Force Estimation}

\begin{figure*}[tp!]
\centering
\subfloat[Force data collection setup]{\includegraphics[width=0.33\columnwidth]{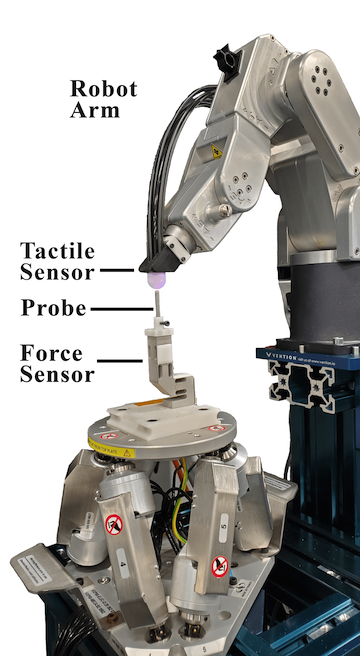}%
\label{fig:characterizationSetup}}
\hfil
\subfloat[Normal force prediction vs. ground truth]{\includegraphics[width=0.4\textwidth]{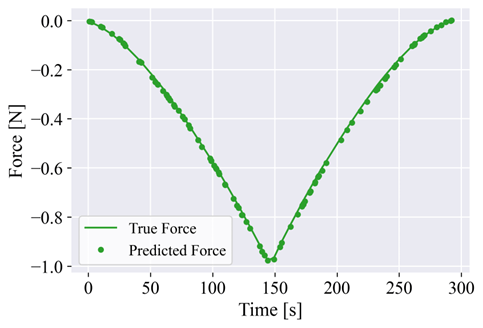}%
\label{fig:normalPrediction}}
\hfil
\subfloat[Shear force prediction vs. ground truth]{\includegraphics[width=0.4\textwidth]{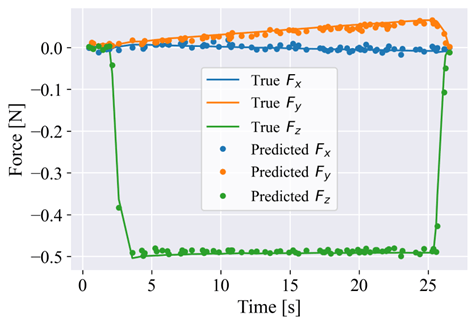}%
\label{fig:shearPrediction}}
\hfil
	\caption{(\subref{fig:characterizationSetup}) Benchmarking setup used to collect force data over three different indenter types. From the dataset, we train the image-to-force prediction models. (\subref{fig:normalPrediction}) We plot the predicted force compared to ground truth for one indentation in the normal force dataset from \SI{0}{\newton} to \SI{-1}{\newton}. (\subref{fig:shearPrediction}) We plot the predicted forces compared to the ground truth for one indentation from the shear force dataset. The robot indents \SI{-0.5}{\newton} in the normal direction (green) before loading in shear (orange).}
\label{fig:forceResults}	
\end{figure*}

\begin{table}[t]
 \small
	\centering
	\renewcommand{\arraystretch}{0.8}
        \caption{Results for Normal and Shear Force Estimation} 
        \label{tab:ForceEstimation}
  
   \begin{tabularx}{\columnwidth}{XXX}
   \toprule
        \textbf{Dataset} & 
        \textbf{Indenter} & \textbf{RMSE~$\pm$~STD~(\SI{}{\milli\newton})}
        \\
        
   \midrule
   \multirow{3}{*}{Normal} & \SI{4}{\milli\meter} & $3.7 \pm 3.2$ \\
    & \SI{12}{\milli\meter} & $3.9 \pm 3.7$ \\
    & Flat & $3.5 \pm 2.9$ \\

    \cmidrule(lr){2-3}
   \multirow{3}{*}{Normal (10k)} & \SI{4}{\milli\meter} & $8.2 \pm 7.4$ \\
    & \SI{12}{\milli\meter} & $6.9 \pm 5.6$ \\
    & Flat & $6.6 \pm 6.5$ \\

    \midrule
    
   \multirow{3}{*}{Shear} & \SI{4}{\milli\meter} & $5.0 \pm 3.3$ \\
   & \SI{12}{\milli\meter} & $3.8 \pm 2.5$ \\
   & Flat & $6.5 \pm 4.0$ \\

    \cmidrule(lr){2-3}
   \multirow{3}{*}{Shear (10k)} & \SI{4}{\milli\meter} & $5.3 \pm 2.6$ \\
    & \SI{12}{\milli\meter} & $6.6 \pm 2.3$ \\
    & Flat & $5.6 \pm 3.0$ \\
   \bottomrule
   \end{tabularx}
\end{table}

We assess force estimation by evaluating \sensor{}'s ability to measure applied normal and shear forces at a contact. Because we are motivated by a tissue palpation use case, or other similar fine manipulation, we are primarily interested in the ability to measure light forces up to \SI{1}{\newton}.

We used an indentor to probe \sensor{}'s sensing element.  As shown in \fig{fig:characterizationSetup}, we mounted the \sensor{} on a \rev{six-axis industrial robot arm (Meca500) with a precision of \SI{5}{\micro\meter}. The metal indenter probe was installed atop the force sensor \rev{such that the sensor frame and probe frame are axis aligned},and the force sensor was tared to the no-contact condition. This assembly was mounted on a hexapod 6-axis stage (Newport HXP50) which can be
precisely controlled to translate with \SI{0.1}{\micro\meter} and rotate with
\SI{0.05}{\degree} increments, but during data collection the stage was stationary and the robot arm translated. 
 More details on the robot indenter data collection setup can be found in Appendix~\ref{app:indentersetup}.} During this process of contacting the \rev{sensor} surface, we collected images of the surface under contact and the contact force vectors from the force sensor embedded under the indenter. We used three different indenter probe tips: \SI{4}{\milli\meter} diameter, \SI{12}{\milli\meter} (on the order of the size of the sensor tip), and a flat metal surface (larger than the sensor tip). The variety in tip sizes produces different contact loading conditions.

Two types of datasets were collected: normal and shear.
\subsubsection{Normal Force Dataset} To characterize normal force performance, we used a single-axis force sensor \rev{(ME-Systems KD34s single axis sensor)} under the indentor probe. For a point on the tip of \sensor{}, the probe pressed perpendicularly into \sensor{} until the normal force of the force sensor crosses a threshold of \SI{1}{\newton}. During contact, the measured normal force and \sensor{} images were collected synchronously for image-to-force calibration at 50 \unit{\hertz} and 10 \unit{\hertz} respectively. The time difference error between the image and force pairs was less than \SI{0.015}{\second}. We downsampled the force data to match the images, and collected approximately 700 (image, force) pairs per spatial point. We repeated this process for 6 random points along a 5×5 \unit{\milli\metre} region of the sensor tip to create our normal force dataset over the course of several days. In total, there are 104,018 image-force pairs. We randomly split these into a training set (80k samples), validation set (10k samples), and test set (remaining 14k samples). We also prepared a much smaller subset of 10,000 randomly-selected image-force pairs with a similar split, which we refer to as ``Normal (10k)." 

\subsubsection{Shear Force Dataset} For shear forces, we used a three-axis force sensor \rev{(FUTEK QMA147 3-axis sensor)} to collect the contact force vector. The procedure is similar to normal force collection at first, with an additional step of loading the probe in shear after normal force is loaded. First, the probe was controlled to apply \SI{500}{\milli\newton} normal force, and then it was moved tangentially to the surface loading shear force up to \SI{80}{\milli\newton}. Finally, the probe was moved back to the previous location, unloading the shear force. If there was non-zero residual shear force, we discarded the data because slip might have occurred. In total, we collected 138,348 image-force pairs over the course of several days to create our shear dataset. We randomly split this into a training set (100k samples), validation set (10k samples), and test set (remaining 28k samples). We also prepared a smaller subset of 10,000 randomly-selected image-force pairs with the same split, which we refer to as ``Shear (10k)."

In training the image-to-force regression models, we used a modified ResNet-18 \cite{he2016deep} architecture deep neural network that takes in an input image of 350×350×3 and outputs a scalar output linear layer predicting the force. One early concern was that fiber-pixelated images would require significant tuning but we found that the ResNet-18 model architecture was sufficient. During training, we used mean square error as the loss and then optimize with Adam with an initial learning rate search. The raw images from the sensor are 640×480 pixels and center-cropped and down-scaled to 350×350 pixels. We found that no other data augmentation techniques were needed for training. We used these specifications to train on the different datasets (Normal, Normal (10k), Shear, and Shear (10k)). For the shear datasets, we also learned to output the force vector direction.

The normal and shear force resolution are reported in \tab{tab:DigitComparison}, and the RMSE values in \tab{tab:ForceEstimation}. We also report a \SI{2.6}{\degree} force-angle precision for the full shear dataset, and a \SI{3.6}{\degree} force-angle precision for the smaller 10k-sample dataset. We report a median absolute error of \SI{5}{\milli\newton} for normal force, and \SI{5}{\milli\newton} for shear force based on the random train/validation/test split. 

\rev{An additional study was conducted on the 10k dataset, where instead of randomly splitting the dataset into the train/validation/test sets, we explicitly use image-force pairs from one spatial point as the validation set and image-force pairs from another spatial point as the test set. This guarantees that the training, validation, and test sets are all from different spatial locations\secondrev{, which are detailed in Appendix~\ref{app:differentspatiallocations}}. The model achieved a median absolute error of \SI{6}{\milli\newton}, a comparable performance with the randomized split.}

Based on our results, \sensor{} is able to accurately estimate light normal and shear forces over the tip of the elastomer surface. There is no significant difference in performance between the different indenter probe sizes, suggesting that \sensor{} would perform well over a variety of contact sizes. The model is also able to learn comparably from both the full dataset and the smaller dataset (10k samples) for this task\rev{, and can learn to predict force on an unseen spatial location at the tip of the elastomer surface.} In the next subsection, we also investigate an analytic model that supports the accurate force estimation performance in the light force regime.

\subsection{Contact Area Model}

Hertzian contact refers to the frictionless contact between two bodies \cite{hertz1881beruhrung}, relating normal force and normal deflection at the contact interface. Given an elastomeric tactile sensor with a hemispherical tip, Hertzian contact theory may inform the expected contact area for contact with simple geometries. Although the elastomer gel is nonlinear, we expect the sensor to roughly correspond to contact mechanics theory when indented in the light force regime. Applicable to the collected indenter dataset are the contacts between a sphere and a sphere, and sphere and half-plane. In this type of contact, the contact spot is a point contact that becomes a contact patch once the bodies deform under loading. The expressions for the circular radius of the contact patch arises from theory, which we summarize briefly below. 

For the case of a sphere contacting a half-plane, the Hertzian contact radius $a$ is given as, 
\begin{equation}
       a = {\left(\frac{3FR}{4E'} \right)}^{(1/3)}\,,
    \label{eq:hertzian_sphere}
\end{equation}
where $R$ is the radius of the indenting sphere, $F$ is the force with which it indents, and $E'$ is the elastic property,
\begin{equation}
\frac{1}{E'} = \frac{1-v_1^2}{E_1} + \frac{1-v_2^2}{E_2}\,,
\end{equation}
where $E_1$ and $E_2$ and the elastic moduli, and $v_1$ and $v_2$ are the Poisson's ratios for each body. For \sensor{}, $E_1 \approx 0.7$\,mPa and $v \approx 0.5$; the indenter is stiffer so the sensor properties dominate.
For a sphere contacting a sphere, the effective radius is 
\begin{equation}
    \frac{1}{R_e} = \frac{1}{R_1} + \frac{1}{R_2}\,,
\end{equation}
and for the case of a sphere contacting a flat indenter, $R_2 \rightarrow \infty$. Thus from Hertzian contact mechanics, we see that contact patch radius is proportional to cube root of the applied load $F$ of the indenting sphere. This expected trend can be verified with image analysis from the training dataset. 

\begin{figure}[t]
\centering
\includegraphics[width=\columnwidth]{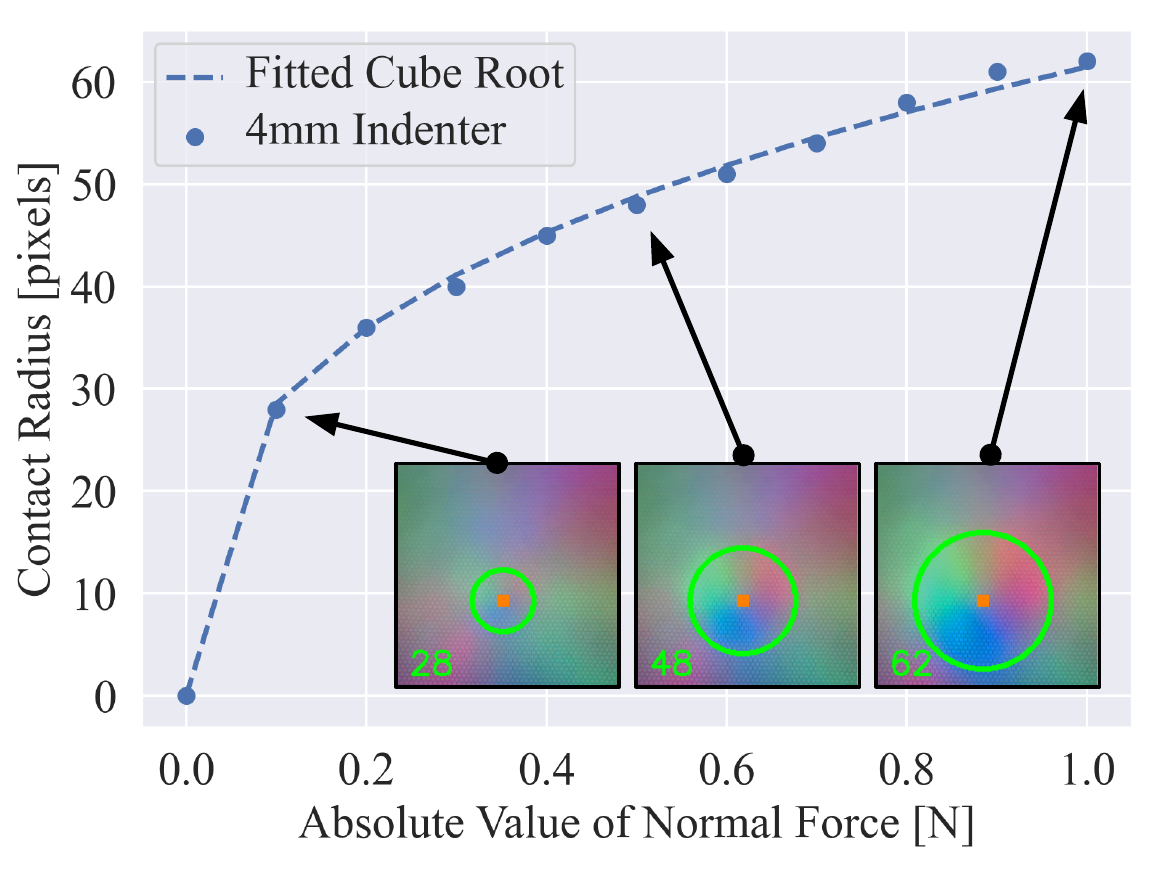}
\caption{Comparison of actual contact area radius and expected trend as the probe indents the sensor surface with increasing force. The dashed line models the cube root fit expected with Hertzian contact. Inset are tactile images with computed radius in pixels.}
\label{fig:contactAreaResults}
\end{figure}

We calculated the contact area radius during an indentation session from $F_z = \SI{0}{\newton}$ to $-1\SI{}{\newton}$ with the \SI{4}{\milli\meter} probe and compare this to the cube root relationship theorized in the Hertzian contact model (\fig{fig:contactAreaResults}). The radius calculation was obtained by subtracting images from the no-contact baseline image, manually thresholding and contouring the resulting difference image, and determining the circle of best fit to the resulting contour. The radius is reported in pixels.

Based on our results, the Hertzian contact model trend fits closely to the observed contact area radius trend for light forces. This suggests that the gel elastic behavior is approximately linear while operating in the light force regime, which aids in learning or analytically solving for contact conditions. We note, however, that this sensor, like other rounded optical tactile sensors, will experience a reduction in sensitivity with higher loads at which the gel deformation becomes less linear and the contact patch grows more slowly. Accordingly, it may be desirable to tune the sensor tip geometry to the expected geometry of an indenting object and to the anticipated load range. 

\section{Silicone Palpation Experiments}
\label{sec:hardness}

Physical examination or palpation of tissue is regularly conducted as part of a preliminary cancer screening. During a physical exam, the clinician should be able to distinguish between the softness of healthy tissue and the comparative stiffness of cancerous tissue.

For a classification task with our sensor, we desire the ability to classify firmness over a range of values that represent both healthy and unhealthy tissue. For this, we prepared a custom dataset associating stiffness (defined as hardness values on the Shore durometer scale for elastomers) to silicone samples of different relevant geometries. Using this dataset, a classification model was trained that classifies the hardness values based on a sequence of 16 image frames, in order to show \sensor{}'s potential for palpation use cases. \rev{An ablative analysis is also provided on the effect of the contact patch and image sequence length to performance.}

\subsection{Hardness Dataset Collection}
To prepare the silicone hardness dataset, custom silicone samples were casted with varied hardness values and geometries by mixing different quantities of Smooth-on Ecoflex 0050 and Smooth-On Smooth-Sil 945 platinum silicones. Ecoflex 0050 (50 Shore 00) was chosen to represent what would be considered healthy tissue, since it is reported to have a similar elastic modulus to real healthy prostate tissues from patients~\cite{navarro2021bio}. Smooth-Sil 945 silicone was chosen because it is significantly harder, has a fast cure time, and has a 1:1 mixing ratio that makes it easy to custom-blend different ratios. The exact mixing ratios used are detailed in the Appendix~\ref{app:hardnessPrep}. 

Molds were prepared for four different types of surfaces: flat, and with bumps of \SI{4}, 
\SI{8} and \SI{12}{\milli\meter} diameter, all raised 
\SI{2}{\milli\meter} above the surface. The bump sizes were chosen based on the different diameter categories reported for lesions in prostate tissues: we desired one bump under \SI{5}{\milli\meter}, one between 5 and \SI{10}{\milli\meter}, and one over \SI{10}{\milli\meter}. The prostate posterior is typically smooth and flat or gently concave, which the flat surface sample mimics. The flat surface sample is also used in a durometer test to find an exact hardness value for each batch. 

\begin{figure}[t]
\centering
\subfloat[Hardness dataset samples]{\includegraphics[width=0.55\columnwidth]{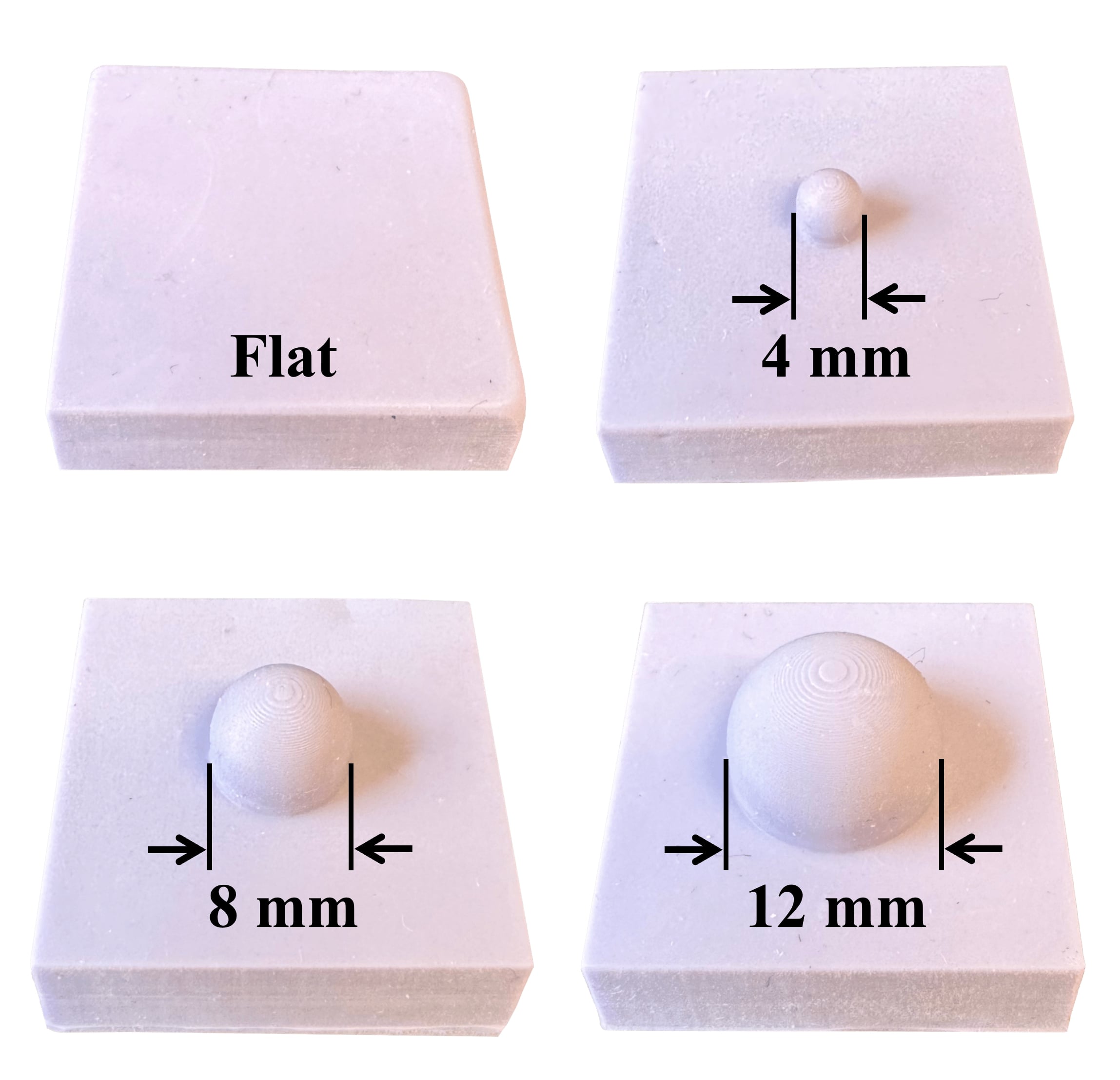}%
\label{fig:hardnessSetupSamples}}
\hfil
\centering
\subfloat[Hardness dataset collection setup.]{\includegraphics[width=0.35\columnwidth]{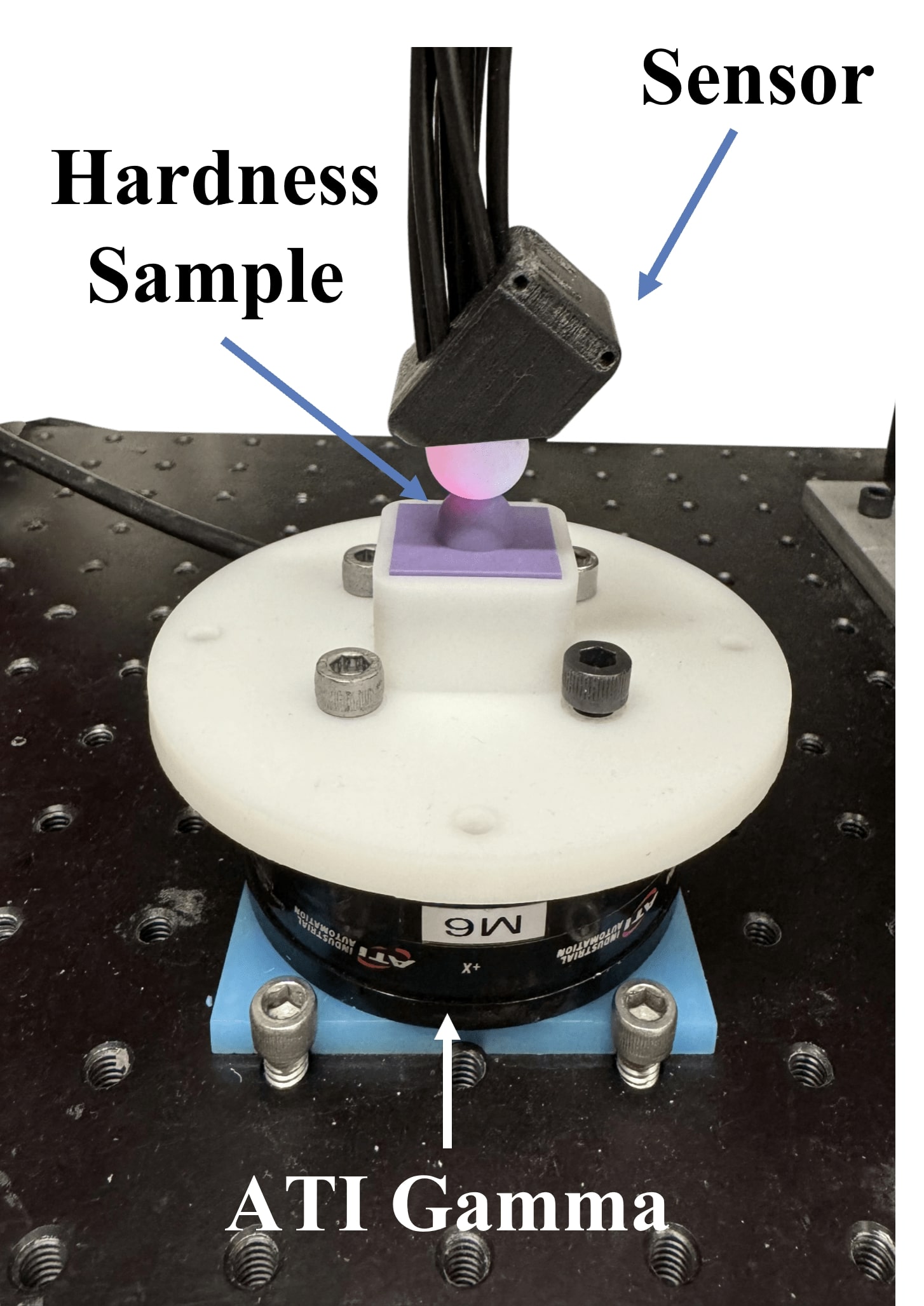}%
\label{fig:hardnessSetupDataCollection}}
\hfil
    \caption{(\subref{fig:hardnessSetupSamples}) Set of silicone samples with the same hardness value for hardness dataset. All sets of samples contain four surface types: flat, \SI{4}{\milli\meter} bump, \SI{8}{\milli\meter} bump, and \SI{12}{\milli\meter} bump. (\subref{fig:hardnessSetupDataCollection}) Experimental setup for collecting silicone hardness dataset. Force data was collected but not used for learning.}
\label{fig:hardnessSetup}
\end{figure}

The molds were printed with a FormLabs 3 printer using Rigid 1K resin and prepared with XTC-3D brush-on coating. The silicone mixtures were then poured into the four different molds to produce four silicone samples per batch. The samples are \SI{7}{\milli\meter} thick and 25 × 25\SI{}{\milli\meter} across; examples are shown in \fig{fig:hardnessSetupSamples}.

In total, 11 sets of silicone samples (44 individual swatches) were prepared in this manner, reflecting 11 different hardness values. The samples were randomly colored with Silc-Pig pigment (0.5\% by weight) to help distinguish between them, but these addition of these colorants did not affect the hardness value. The hardness values were assigned to each sample batch by taking the average of five measurements with the durometer on the flat surface sample, where the 0.5A sample is the healthy tissue baseline. 

The experimental setup shown in \fig{fig:hardnessSetupDataCollection} was used to manually palpate the silicone samples for the dataset. As shown in the figure, the samples were mounted on a 3D-printed holder that was rigidly bolted to an ATI Gamma force-torque sensor. During the palpation session, a computer synchronously recorded both image and force data.

The palpation technique we used for the dataset collection was to manually make and break contact with the sensor tip normal to the surface of the silicone feature. We gently palpated (no more than about \SI{2.5}{\newton} normal force) each silicone sample feature three times for one minute each, mimicking the gentleness necessary for patient comfort in a real examination. Each palpation session consisted of the raw image sequence from the tactile sensor and the forces from the ATI Gamma. The force data was not used during model training.  

\subsection{Classification Results} 

\begin{figure}[t]
\centering
\includegraphics[width=\columnwidth]{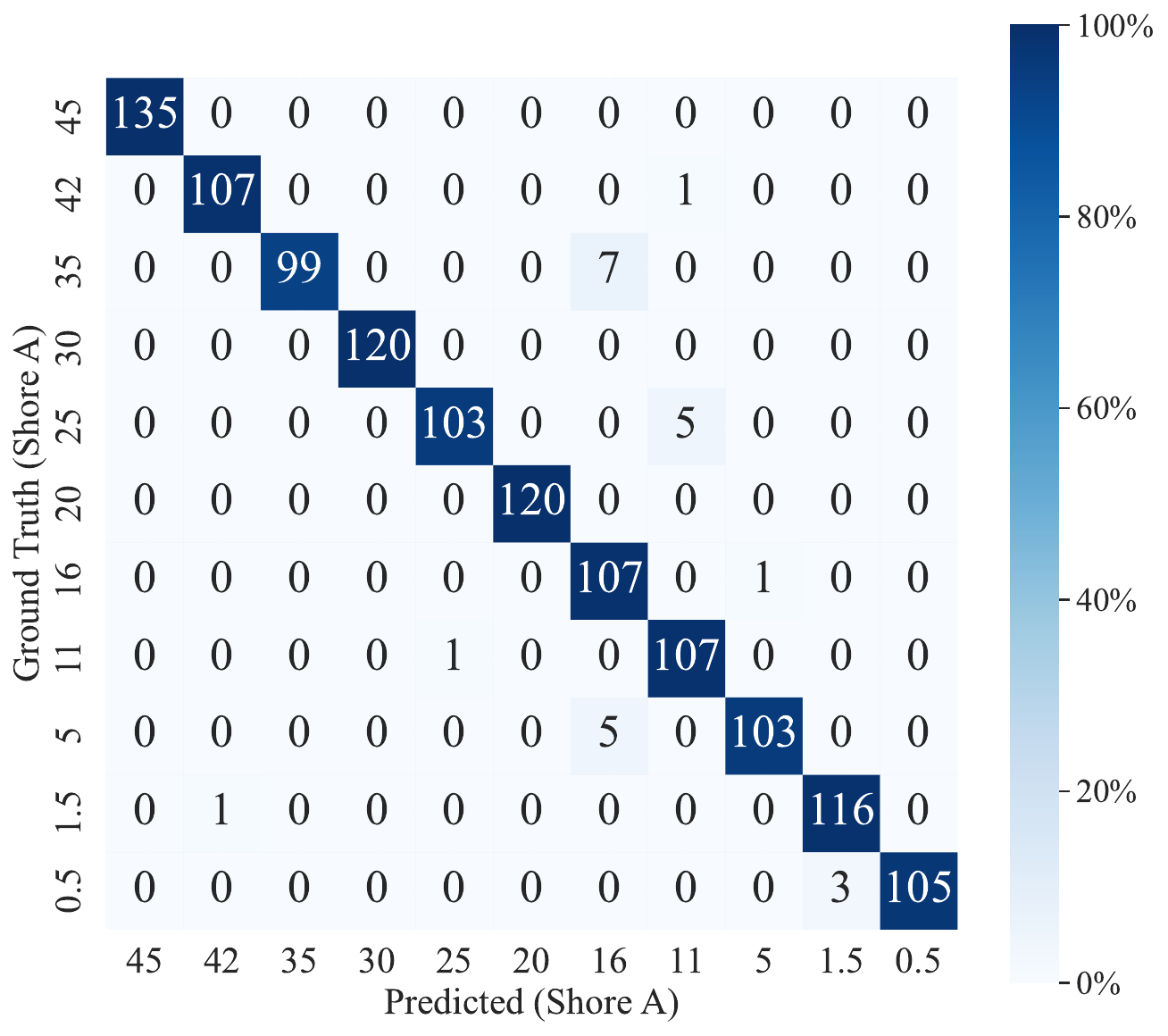}%
 \caption{Classifier performance on silicone hardness class prediction task. Classification results on the silicone hardness samples show that the learned model is capable of predicting the hardness of the samples being touched with high accuracy.}
\label{fig:rubberRubberClasspalpationResults}	
\end{figure}

To train the classification model, we first prepared the dataset---the raw RGB frames are cropped and downscaled to 224 × 224 pixels. To create the training, validation, and test sets we dedicated from each video frame sequence the first 80\% of the frames for training and the last 20\% of the frames for validation and testing. From the initial 80\% of the video sequence, the training set was created by sampling from overlapping sequences of 16 frames with stride 5. Validation and test sets were created from the remaining 20\% of the video sequence by sampling non-overlapping sequences of 16 frames, where the first 10\% is for validation and the remainder for test.

To perform classification, we fine-tuned transformer-based video masked autoencoders (VideoMAE) \cite{tong2022videomae}. Fine-tuning was performed by training of a linear model on base model representations. 
The model takes as input the 16 frames sequences, and outputs a scalar that represents the hardness class. The models were implemented with PyTorch and the large instance of a pre-trained VideoMAE model openly available on HuggingFace.
We used Adam optimizer with learning rate 0.005 and batch size 8, and a cross entropy loss for the classification task. 
The results are shown in the full confusion matrix (\fig{fig:rubberRubberClasspalpationResults}). 

After evaluating the trained classification model on the test set, the model was capable of reaching \rev{98\%} accuracy over \rev{11} hardness classes. Based on the results, the trained model performs well in classifying elastomer hardness values that correspond to the range from healthy to unhealthy tissue.

\subsection{\rev{Analysis of Contact Surface Properties and Image Sequence Length for Hardness Classification}
}
\label{subsec:ablation}

\rev{To further study the learning ability of the VideoMAE model, rather than using the tactile images, delta-frames were instead used as input, computed as the difference between the current frame and the next frame (i.e.  $\Delta_{l+1,l} = \text{frame}_{l+1} - \text{frame}_l$). Using delta-frames ensured that the model was learning mostly from the evolution of the contact patch and not spurious artifacts. Training on the 16-frame sequence of delta-frames yielded an accuracy of 94.4\% over 11 hardness classes, comparable in performance to that from training with the tactile images. This result gave confidence that the model was learning to classify hardness based on the contact patch, rather than obscure artifacts not grounded in physical properties of the contact.}

\begin{figure}[t]
\centering
\subfloat[\SI{12}{\milli\meter} diameter bump made of stiff material]
{\includegraphics[width=0.47\columnwidth]{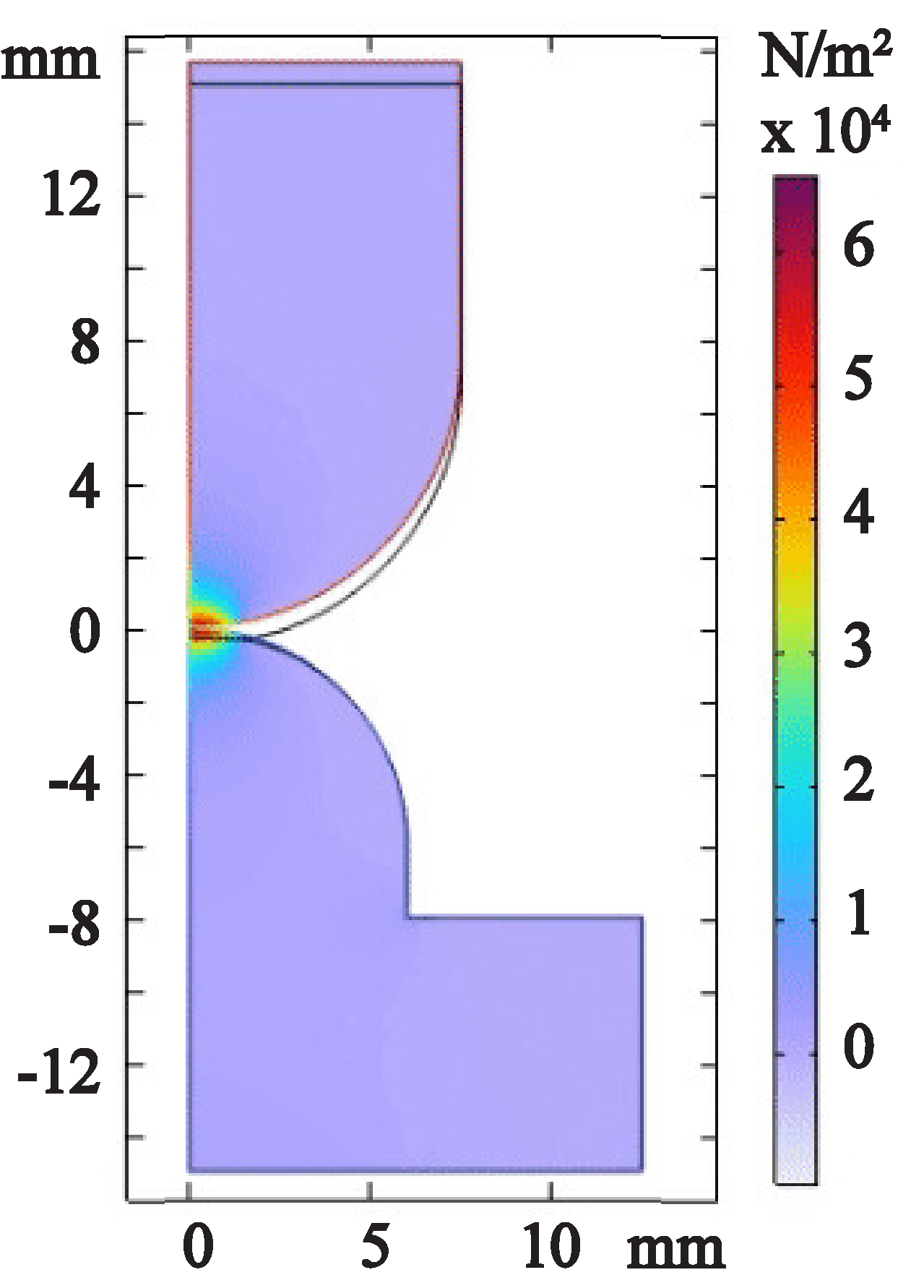}%
\label{fig:12mmFEA}}
\hfil
\centering
\subfloat[\SI{4}{\milli\meter} diameter bump made of pliable material]{\includegraphics[width=0.5\columnwidth]{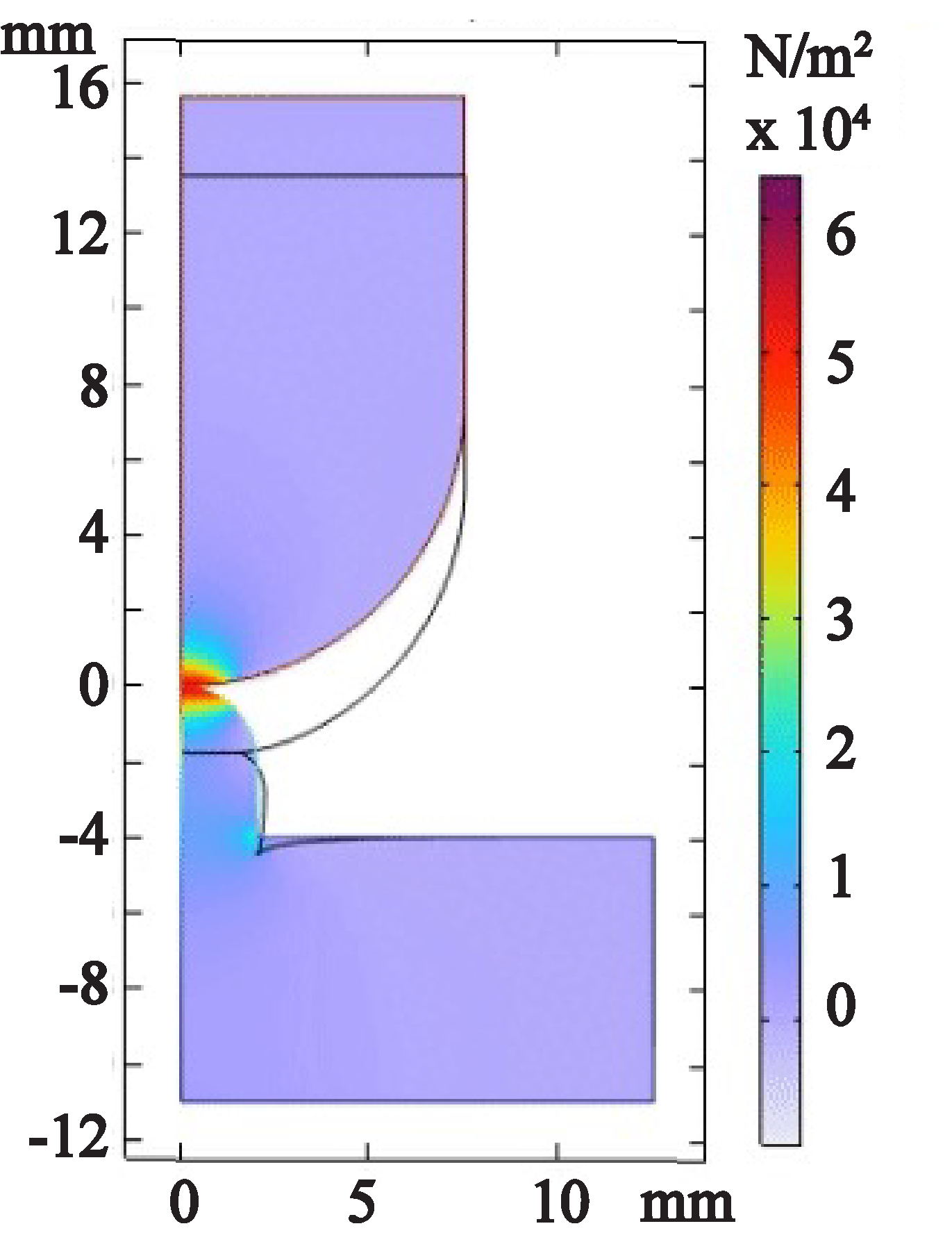}%
\label{fig:4mmFEA}}
\hfil
    \caption{\rev{FEA models of potentially confounding indentation scenarios. Shown are the deformation magnitude and pressure distribution for (\subref{fig:12mmFEA}) \SI{12}{\milli\meter} diameter bump made of Shore 15A material, and (\subref{fig:4mmFEA}) \SI{4}{\milli\meter} diameter bump made of Shore 0A material. Black lines depict the deformation of the materials in each case. Coloring depicts the surface pressure distributions. At \SI{0.5}{\newton}, they have the same contact radius but different pressure distributions.}
    }
\label{fig:FEA}
\end{figure}

\rev{We note, however, that perceived material stiffness is a function of both material hardness \emph{and} geometry, and yet the classification results were quite accurate despite the different diameter geometries in the sample dataset. Although the evolution of the contact patch was important, it was likely not the only discriminant to the machine learning model. To show this, we highlight an adversarial case: a sample with a large diameter bump made of a stiff material and a sample with a small diameter bump made of pliable material could have the same contact patch size when indented. This scenario was represented in the collected silicone hardness dataset, and was modeled using COMSOL Multiphysics, a finite element analyzer. Because the scenario is radially symmetric, only half of the indenter and sample were modeled for computation speed, with the model parameters given in Appendix~\ref{app:FEA}. As shown in \cref{fig:FEA}, although the contact patch radius was the same, the different bump diameters lead to different surface pressure distributions when indented with a \SI{0.5}{\newton} load. Even in adversarial cases where the sensor may image a similar contact patch, there was sufficient resolution to capture other physical properties of the contact, explaining in part how the model discriminates hardness classes accurately despite different sample geometries.}

\secondrev{We also note that in the ideal case, indenters should be matched in mechanical impedance with their samples for optimal sensitivity~\cite{yuan2017shape, yuan2016estimating}. In this work, the Shore 10A indenter material (around Shore 00-60) was 10 degrees harder than the softest samples in the dataset (Shore 00-50). There was therefore a reduction in sensor sensitivity because \sensor{} used a stiffer-than-optimal material for the softest samples. However, the material mismatch was not severe enough to significantly degrade performance on soft samples. We provide more analysis in Appendix~\ref{app:stiffnessminimumforce} to explore the tradeoff between indenter material and sensitivity.}

\rev{A study was also conducted on the image sequence length. In this study, the model used the full 16-frame sequence of delta-frames as input, where the $n$ last frames in the sequence were masked by black frames for $n = 0, 8, 12, 14, 15$. This corresponded to effective image sequences of 16, 8, 4, 2, and 1 delta-frames respectively. The F1 score results are plotted in \cref{fig:rubberMaskResults} and the accuracy scores in Appendix~\ref{app:ablation}. Based on the results, the VideoMAE model performed best with longer image sequences, suggesting that sequence length is important for task performance.} 

\begin{figure}[t]
\centering
\includegraphics[width=\columnwidth]{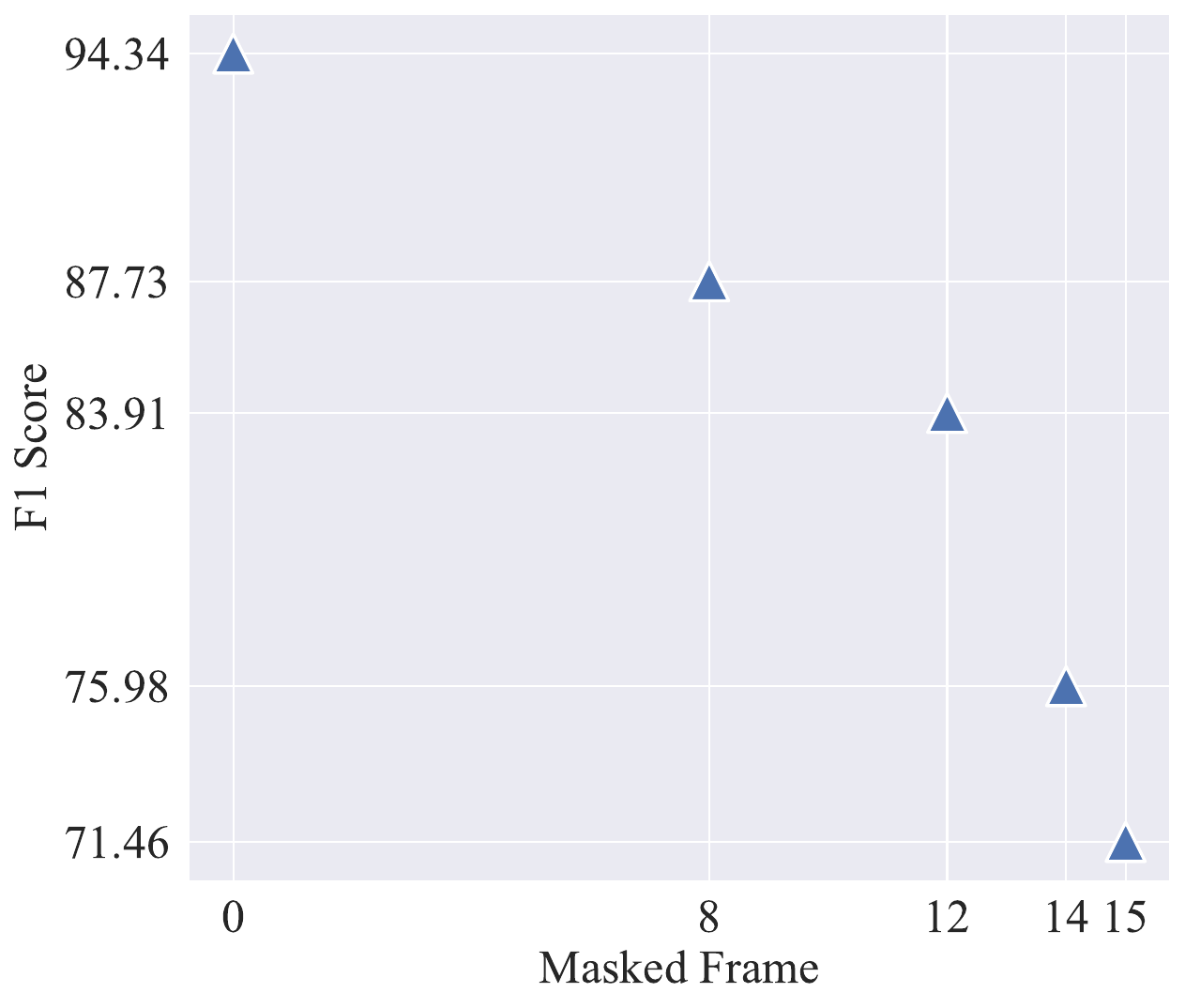}%
 \caption{\rev{The F1 scores for the image sequence length study on the silicone hardness classification task with delta-frames as input ($\Delta_{l+1,l} = \text{frame}_{l+1} - \text{frame}_l$). Using all 16 frames, with no masked frames, led to the highest F1 score. F1 scores decreased as more delta-frames were masked in the sequence, showing that shorter sequences worsened classification performance. }}
\label{fig:rubberMaskResults}	
\end{figure}


\section{Prostate Palpation Case Study}
\label{sec:casestudy}

\sensor{} provides high resolution tactile sensing that could be used to determine tissue hardness in a DRE or similar examination that would be difficult to conduct without a 
human finger-sized tactile sensor. We tested \sensor{} in a clinically relevant task by performing palpation on phantom prostate tissue and on an \textit{ex vivo} tissue specimen. 

\subsection{Classification of cancerous lesions in phantom prostate}

\begin{figure}[t]
\centering
    \includegraphics[width=\columnwidth]{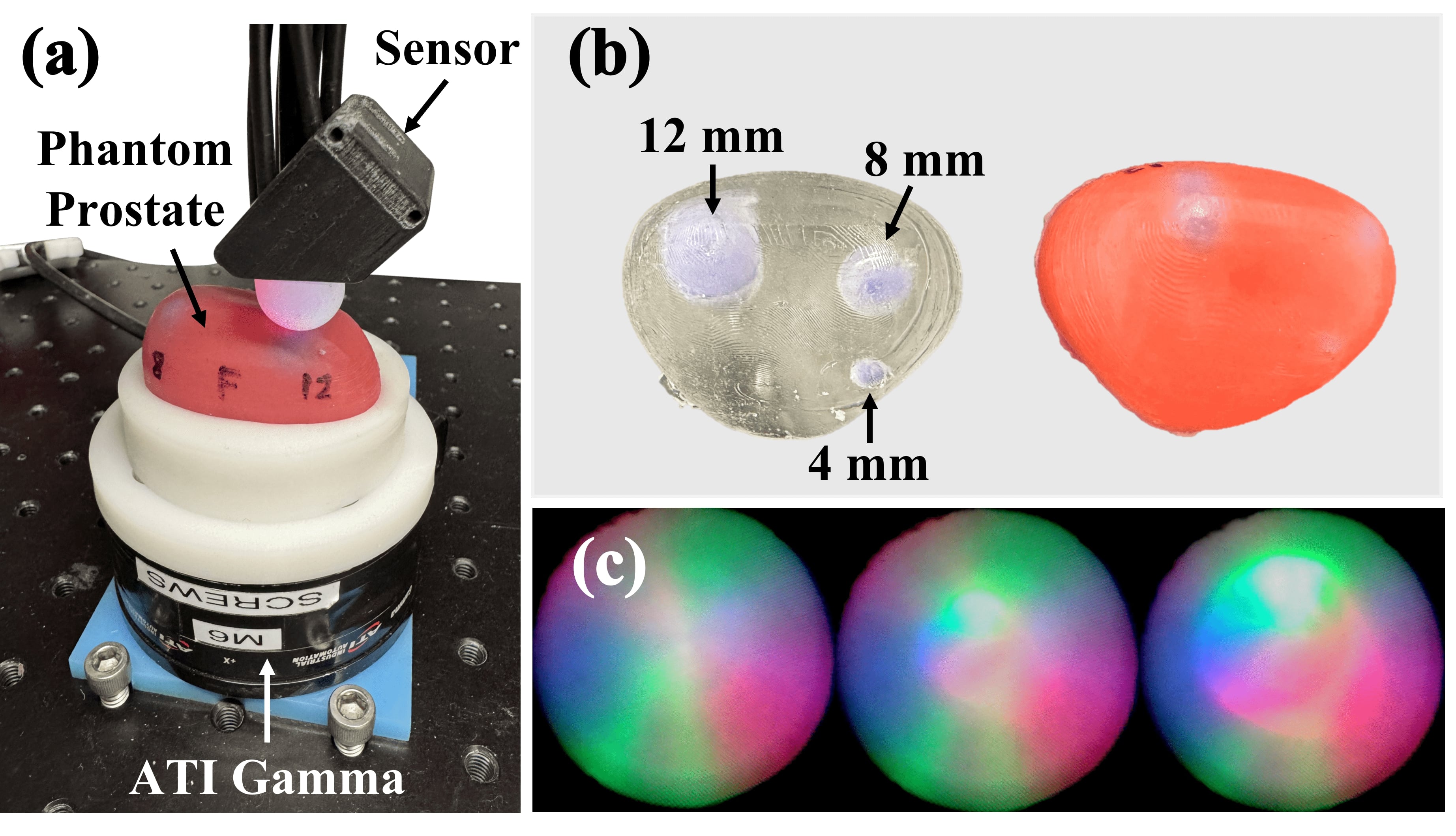}%
    \hfil
	\caption{(a) Experimental setup for collecting phantom prostate hardness dataset. Force data was collected but not used for training. (b) \textit{(Left)} A transparent phantom made from medical gel to clearly show the embedded nodules. \textit{(Right)} An example of the prostate phantoms used for data collection. (c) Frames from indentation sequence on a \SI{12}{\milli\meter} embedded lesion using the ungloved \sensor{} tip.}
\label{fig:phantomSetup}
\end{figure}

\begin{figure}[tp!]
\centering
    \includegraphics[width=\columnwidth]{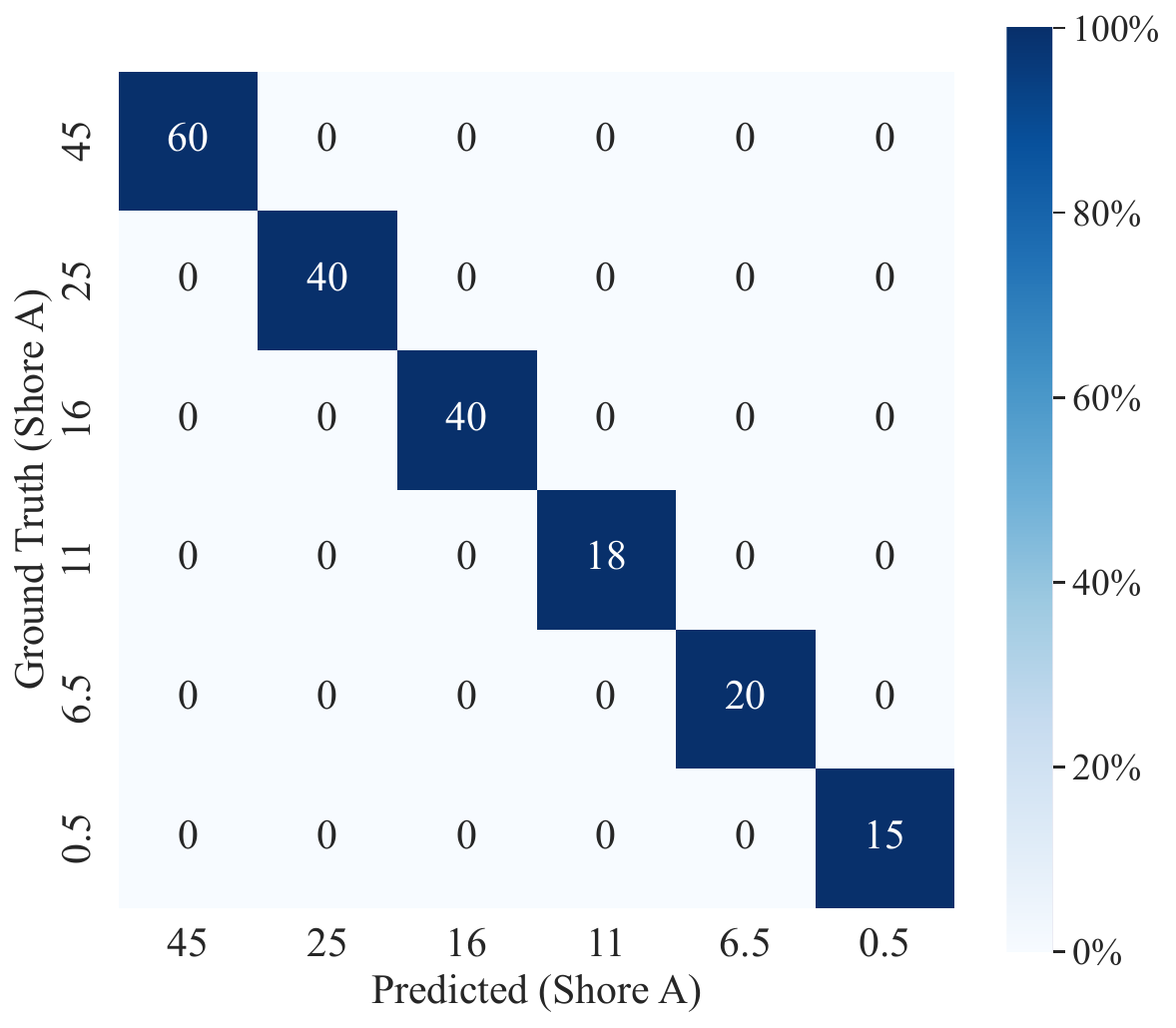}%
    \caption{Classifier performance on phantom hardness class prediction task. Prediction results on the phantom hardness samples show that the model can accurately distinguish between 6 different types of nodules embedded inside the phantom.}
    \label{fig:phantomPhantomClasspalpationResults}
\end{figure}

We desired to test if \sensor{} could make clinically relevant distinctions between cancerous lesions and healthy tissue in a prostate phantom. 
Six phantom prostates were first prepared by creating a mold from a patient prostate model, open-sourced by the Johns Hopkins Urology Robotics Program, NCI 5R01CA247959. The \SI{41}{\centi\meter}$^2$ prostate model was selected, as this was the closest size corresponding to the average volume over four radical prostatectomies that were witnessed at Stanford Hospital. A mold was created only from the posterior half, which is the portion that would be palpated in a digital rectal exam. The mold is 3D-printed using a FormLabs3 printer with Rigid 1K resin and prepared with XTC-3D brush-on finish \rev{to ensure that the surface was smooth}.

We cast the healthy portion of the posterior prostate using Ecoflex 0050 (50 Shore 00), which was chosen for having a similar elastic modulus to healthy prostate from patients~\cite{navarro2021bio}, producing a 0.5A hardness value with a handheld durometer. We colored the Ecoflex 0050 silicone pink using 0.5\% Silc-Pig red and 0.25\% Silc-Pig white by weight. During the curing process, we then embedded harder spherical nodules of \SI{4}{\milli\meter}, \SI{8}{\milli\meter}, and \SI{12}{\milli\meter} inside the phantom at a depth of \SI{2}{\milli\meter} to mimic a cancerous lesion. The mixtures for the spherical nodules were mixed with the same ratios detailed in the Appendix~\ref{app:hardnessPrep}. The phantom prostates are fully cured at room temperature after 4 hours. 

We chose to embed five different nodule hardness values inside the phantom prostates (45A, 25A, 16A, 11A, and 6.5A) to capture a range of lesion instances. As in the previously described experiment, we manually verified the hardness values of the nodules by taking the average of five measurements with a durometer from a flat silicone piece cast in the same batch as the nodule. As for the choice of hardness value for the nodules, a previous study on DRE-detectable nodules used hardness values of 23A, 27A, and 31A in simulated tissue~\cite{baumgart2010characterizing}, but in consultation with clinicians at Stanford Hospital who regularly perform the DRE on patients, it was apparent that prostate hardness values for healthy and unhealthy tissues vary widely from patient to patient. Therefore, it was desireable to make phantoms that reflected a wider range of possible lesion hardness values. The verisimilitude of the final phantom prostates was qualitatively assessed with clinician feedback. An example of one of the prostate phantoms is pictured in \fig{fig:phantomSetup}.

The classification experiment on the phantom tissues was conducted using data from the phantom prostate. We collected the phantom prostate data by manually palpating with \sensor{} on each phantom, which was mounted atop the ATI Gamma force-torque sensor, as shown in \fig{fig:phantomSetup}. We gently palpated (with no more than \SI{2.5}{\newton} force) over healthy phantom tissue (without an embedded nodule), and directly over each embedded nodule size. We palpated each point three times for one minute each with the usual \sensor{} tip and also a tip wearing a \SI{5}{\milli\meter} thick nitrile fingertip cut from the pinky finger of a small medical nitrile glove. The nitrile glove was added to simulate the nitrile glove that a clinician would wear when conducting the exam. Each session was recorded with synchronized raw video and force data although, as with the previous experiment, the forces were not used in model training.

The classifier model and parameters were set up as described in the previous experiment in Section \ref{sec:hardness}. We trained the classifier on the phantom palpation dataset, which included all embedded nodule sizes (4, 8, and \SI{12}{\milli\meter} diameter), the six detected hardness classes (where 0.5A is the baseline healthy tissue), and palpations with a gloved sensor and an ungloved sensor. 

As shown in \fig{fig:phantomPhantomClasspalpationResults}, we achieved a perfect \rev{(100\%)} accuracy score on the classification task, suggesting good performance in hardness classification on phantom prostates. We note that the classification worked well even when the sensor was gloved. Furthermore, the classification task is able to achieve respectable performance even given the different simulated lesion sizes. A human clinician might have trouble feeling the smallest \SI{4}{\milli\meter} nodule during a real exam~\cite{baumgart2010characterizing}. 

\rev{For the phantom prostate data, we also repeated the studies on contact patch evolution and the image sequence length using the same delta-frame input preparation as described in \cref{subsec:ablation}. Training on a 16-frame sequence of delta-frames yielded an accuracy of 97.97\%, giving confidence that the model was learning from the physical properties of the contact with the phantom prostate.}

\rev{Similarly, the effects of image sequence on performance were evaluated with sequences of $n$ masked frames where $n = 0, 8, 12, 14, 15$, corresponding to 16, 8, 4, 2, and 1 delta-frame sequences. The F1 scores and accuracy scores are also included in Appendix~\ref{app:ablation}. As with the silicone hardness samples, the model performed best on the phantom prostate samples with longer image sequences. }

\subsection{Classification of cancerous lesion in ex vivo prostate tissue}

\begin{figure}[t]
\centering
	\includegraphics[width=1.0\columnwidth]{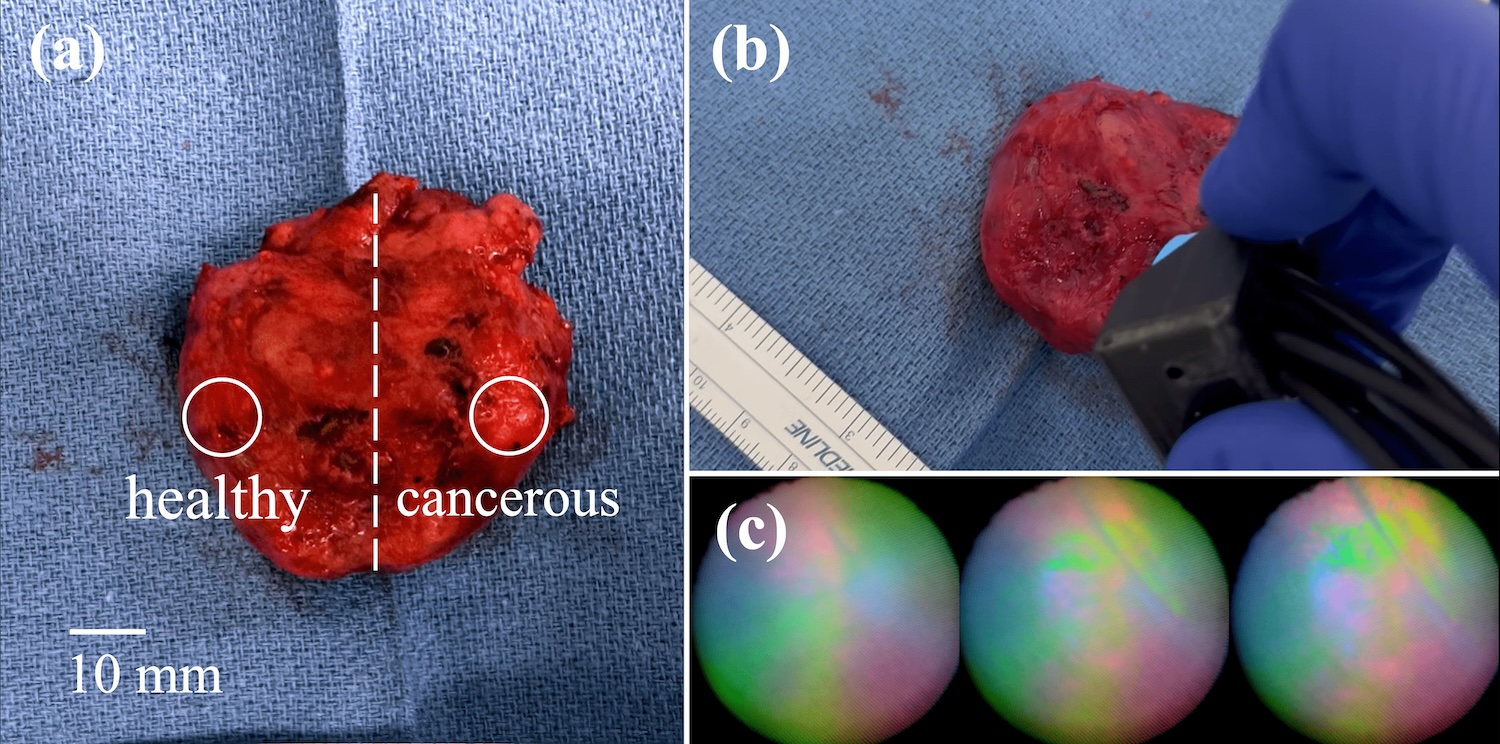}
	\caption{(a) Photo of the \textit{ex vivo} prostate immediately after a radical prostatectomy procedure. Highlighted are the two points palpated on the posterior: \rev{the left point outlines the non-cancerous, healthy region and the right point outlines the cancerous region}. (b) Photo of manual palpation of \rev{the cancerous point} with a gloved sensor tip. (c) Frames from an indentation sequence of \rev{the cancerous point}. The visible wrinkle is from the glove \rev{that is put on the sensor tip}.
    }
	\label{fig:exVivoProstate}
\end{figure}

We brought the \sensor{} to the operating room to manually palpate a resected human prostate gland immediately after a radical prostatectomy procedure. No distinction was made between tumor size, depth, or Gleason Grade of cancer when choosing to shadow this procedure. Because we palpated a fresh specimen, we mitigated the effects of the mechanical property changes that soft tissues naturally undergo when excised. The resected prostate gland is shown in \fig{fig:exVivoProstate}. 

This biospecimen had a Gleason Grade of 7 and was asymmetrical, where the right side contained a cancerous lesion and the left side was healthy. The attending surgeon qualitatively deemed the lesion as nonpalpable by DRE.

With a gloved \sensor{} tip, the specimen was manually palpated (making and breaking contact) approximately thirty times on the posterior surface over a point on the right section (cancerous), and approximately thirty times over a symmetrical point on the left section (healthy). Data were collected over a course of six minutes. Because we had a limited amount of time with the specimen, we did not sample more points spatially. 

We also manually palpated the same areas with a handheld portable durometer (Model: WonVon QMLBH0730HA-C-D). The durometer data for the point on the right side (cancerous) averaged to Shore 7.5A, and the point on the left side (non-cancerous) averaged to 5A. Because this is a handheld durometer on a specimen surface that is not entirely flat, these datapoints should only be used as a reference and not ground truth values. However, the values suggest that the hardness difference between the two points is small, reaffirming the attending surgeon's judgement that the tumor was nonpalpable at DRE.

\begin{figure}
\centering
    \includegraphics[width=0.9\columnwidth]{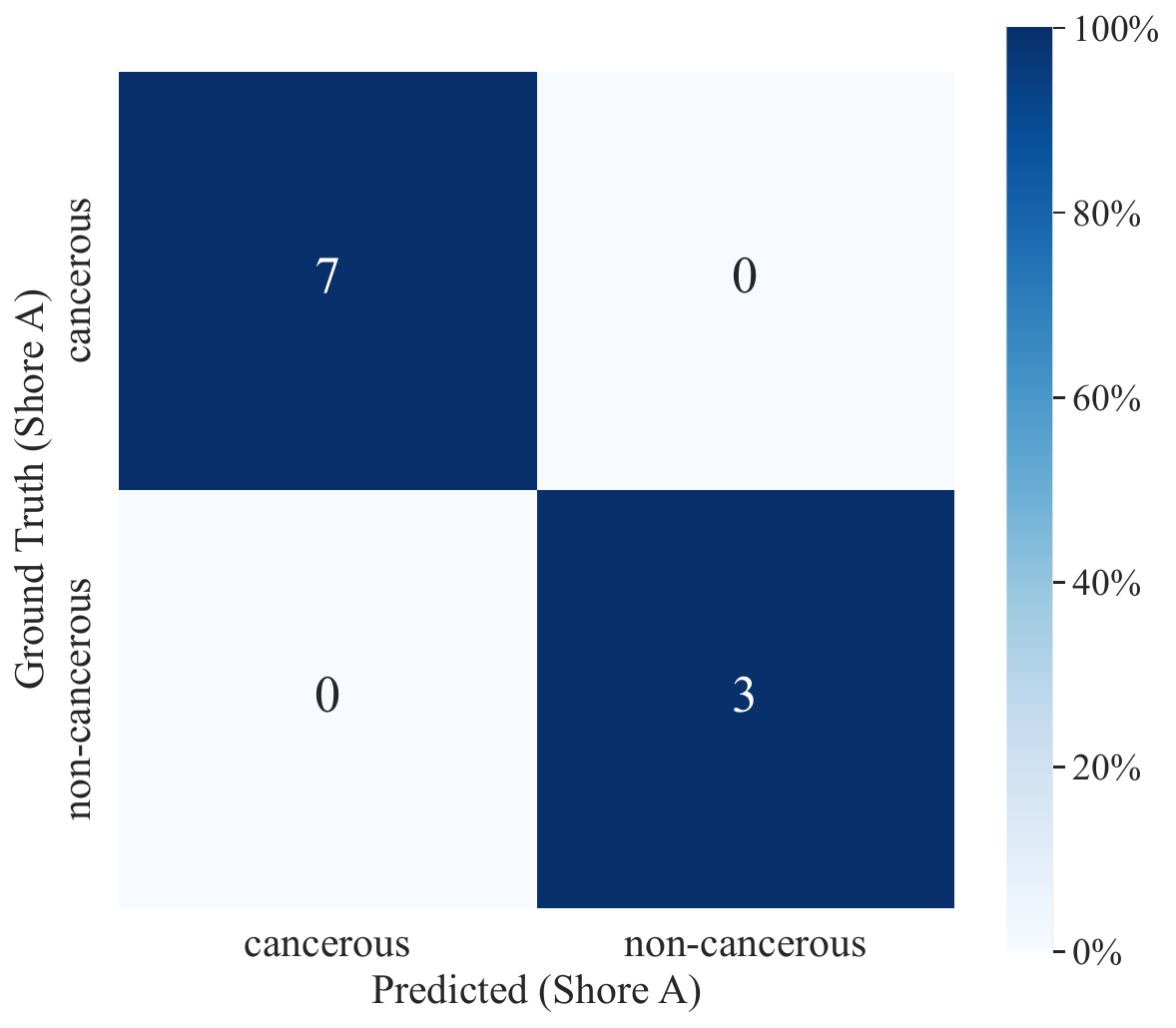}

    \caption{Classification performance on \textit{ex vivo} prostate tissue. The experimental outcomes show that the trained model distinguished between non-cancerous and cancerous tissue in the test set.}
	\label{fig:exvivoExvivoClasspalpationResults}
\end{figure}

We trained the model, as described in \sec{sec:hardness}, on the binary classification task. This classification experiment differs from the previous one only in the creation of training, validation and test sets. Due to the small dataset, we dedicated one third of the videos for constructing the training set, one third for the validation set, and the remaining third for the test set. We created a training set by sampling overlapping sequences of 16 frames with stride 2. Validation and test sets were created by sampling non-overlapping sequences of 16 frames.

We measured perfect accuracy on the test set, with \fig{fig:exvivoExvivoClasspalpationResults} showing the confusion matrix. \secondrev{Despite the more complex geometry of the prostate, it was not entirely a surprise that \sensor{} was able to perform well. Previous literature has shown that vision-based tactile sensors perform better at hardness estimation on smooth and simple rather than sharp geometries~\cite{yuan2016estimating,yuan2017shape}. Prostate glands are smooth bulk tissues with no sharp edges (the dark-colored blood and tissue on the surface can be visually confounding and may give the appearance of complicated geometries in photographs), so they deform predictably.} Because we were only able to sample one real prostate specimen, the classification result here \rev{did not represent a full medical study}. However, the result suggests that the tactile sensor is able to distinguish between the two sampled points with different stiffness values and supports the suitability and clinical relevance of \sensor{} on real tissue.


\section{Discussion and Future Work}
\label{sec:discussion and future work}

In this work, we have demonstrated a new tactile sensor design that enables miniaturization 
beyond that achieved by most vision-based tactile sensors. The work is motivated by a desire to create fingertips with dimensions no larger than those of human fingertips, while maintaining the spatial and force resolution that have made vision-based tactile sensing popular. The resulting sensor is suitable for manipulation and exploration tasks where space is constrained, including the palpation of human organs. The experimental results demonstrate that \sensor{} can learn clinically-relevant information from a representative palpation task, and provides merit for learning the details of the contact patch as part of diagnostic procedure.

\subsection{\rev{Limitations}}

The proposed design is not without limitations. A major limitation is cost; manufacturing a small hyperfisheye lens prototype in limited quantities is expensive, so only the tip of the sensor could be imaged. The cost, however, would decrease when produced in bulk. Furthermore, finished coherent imaging fiber bundles and optical imaging lens pieces are also expensive, though raw fiber spools may be purchased at a cheaper rate.

Another limitation of this design is that miniaturization generally sacrifices resolution. In this design, because the fiber cores are the pixels of an image, the quality and quantity of the fiber cores dictate the resulting resolution. Although \sensor{} has enough spatial resolution to perform the palpation task accurately, it could have difficulty performing other manipulation tasks that require much higher resolution, such as very fine-grained texture classification. A broader exploration of how much resolution is needed for certain tasks is out of scope, but warrants further investigation.

Finally, fiber bending must be considered. The fibers themselves have a minimum bend radius (to ensure no breakage or light loss). It is reasonable to think that this sensor design as is would not suitable for high-impact forces or for settings where the fibers may experience sharp bends beyond the minimum bend radius, which generally scales inversely with cost. For context, in this work, the minimum bend radius was \SI{55}{\milli\meter} (but this was not optimized during the design process because clinicians do not bend their fingers during rectal examinations). It is relatively straightforward to buy affordable fiber bundles (a few U.S. dollars per 10 ft.) with minimum bend radii down to \SI{15}{\milli\meter}, but beyond that they could become costly.

This work also did not investigate fatigue or cycle testing with repeated bending of fibers, which we leave as future work. However, extensive mechanical fatigue testing of optical fibers has been conducted in the literature to millions of cycles~\cite{glaesemann2017optical}, demonstrating better fatigue performance than copper wire common in prototyping~\cite{donald1967stranded}. Defects in the silica can degrade and propagate if the fibers become wet, however almost all commercially-available fibers (including those used in this work) are sheathed with a protective jacket to prevent degradation from moisture and abrasion.

\subsection{Future Design Directions}

Ample further opportunity exists for tuning a fiber-based approach specifically to tactile sensing. As noted earlier, it would be desireable to substitute the lens in this prototype with a hyperfisheye lens, assuming reasonable cost, to image the whole surface of the sensor. Details on the considerations for designing a hyperfisheye lens specifically for tactile sensing are noted in \app{app:hyperfisheye}. Another direction includes researching different fiber and microlens architectures, taking advantage of the optical customization that fibers offer. An immediate extension to increase the fiber core count, either through purchase or by manually heating, stretching, and combining discrete fibers.  

However, one does not necessarily need to use a standard lens to achieve a short working distance at the distal end. An interesting design direction is to replace the lens with a microlens array, producing an optical configuration akin to the compound eyes commonly found in arthropods. 

Changes to the choice of imaging conduit could also be investigated. Because this work assumed that the resulting tactile images needed to be legible to humans, a coherent imaging fiber bundle was chosen. However, this is not a strict requirement for a tactile sensor, especially given the capabilities of machine learning. One may instead use a non-coherent fiber wave guide for imaging, which tends to be cheaper than coherent imaging bundles, but could increase the computation burden when training.

In addition, \sensor{} could be pushed further in terms of miniaturization. For example, if all the fibers in this initial prototype were combined in a singular bundle, this design could be \SI{8}{\milli\meter} in diameter or smaller, with potential uses in endoscopy. Other experiments could also tune the camera's dynamic range, exposure, and lighting to better capture details or reduce hotspots.  

As became clear with the tested prototype, there is a trade-off between small size and resolution. One current drawback with a conventional fiber-based imaging system is that the number of pixels in an imaging detector (e.g., a remote CCD) is typically much greater than the number of fibers in a fiber bundle, so fiber-based imaging tends to be less sharp than a direct camera image. A related drawback is that images transmitted through fibers undergo a hexagonal sampling from the fiber packaging followed by a rectangular sampling through the imaging detector. The mismatch between essentially two sequential sampling operations leads to some data loss. To account for this, one could chose to increase the core count or use deep learning approaches to upscale or generatively fill in images.
Alternatively, if an extremely small diameter is not a design priority, one can investigate a hybrid of fiber-based lighting (allowing customizable optics and mitigating heat effects at the distal end) with an endoscope ``chip-on-tip" CCD imager.

\subsection{Additional Applications}

\begin{figure}[t]
\centering 
    \includegraphics[width=\columnwidth]{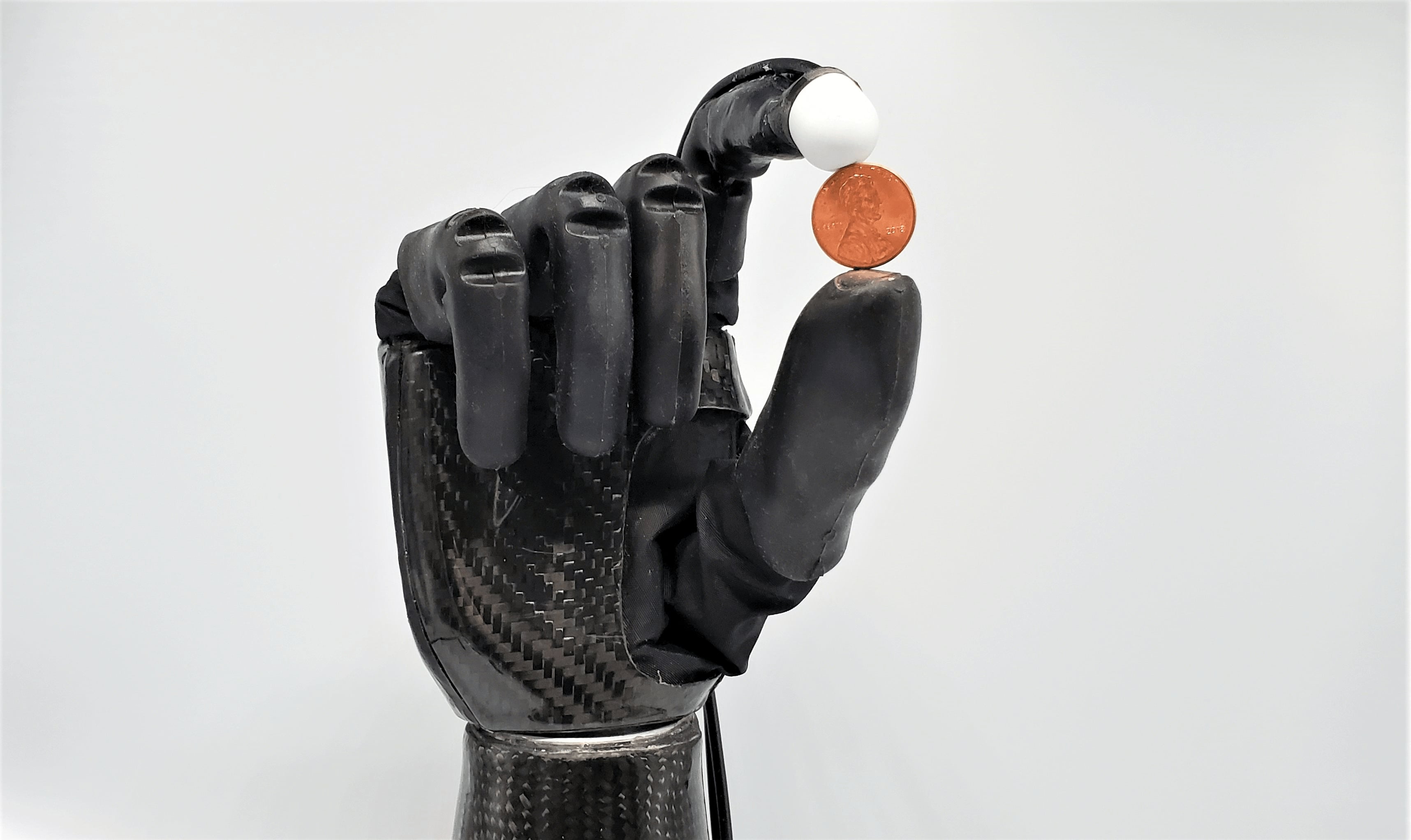}
	\caption{A concept photo showing that the compact design of \sensor{} is sufficient for integration as the fingertip of a Psyonic Ability hand (holding a U.S. penny for size comparison).}
	\label{fig:demo}
\end{figure}

Looking further ahead, we can exploit other benefits of a fiber-based approach, such as information transfer over long distances and resistance to electromagnetic radiation, for new applications in rugged environments like in an MRI machine, underwater or in outer space. In this case, it will be necessary to retain the remote-electronics aspect of the present design.

Within the realm of medical applications, we would like to investigate other patient examinations that require understanding the tactile properties of tissue. For example, cervical tissue stiffness is an important indicator for preterm births and this sensor could be used to palpate cervical tissue to characterize stiffness. Similarly, during an open surgery of the pancreas, the surgeon will often palpate the pancreas because texture is an important predictive factor of post-operative complications. A robotic counterpart could provide repeatable quantitative measurements of texture by using a sensor like \sensor{}. One may modify \sensor{}'s gel to be selectively transparent or opaque, for applications where one may need standard imaging in addition to tactile imaging in constrained settings.

A fiber-based approach is promising beyond medical applications as well. A sensitive fingertip could be integrated with a human-scale dexterous hand (\fig{fig:demo}) for tactile exploration and manipulation of delicate items. Future work could take advantage of the spatial resolution to discern object textures and forces for tasks like ripe fruit handling or cloth folding. 

The approach of the \sensor{} design to separate the sensing element from supporting circuitry could also be expanded upon for applications with multiple fingertips or even for robotic skins. A future dexterous robot hand with multiple sensorized fingertips, in lieu of using multiple traditional vision-based tactile sensors (requiring a camera, USB cable, and associated wiring management per finger), could instead have multiple passive gel tips combine to a single optical fiber bundle, camera, and USB connection downstream. A fully sensorized robot hand with tactile fingertips and a large-area skin (for the palm or back of the hand), could also be made possible with multiple optical fiber bundles. Although routing a fiber bundle may prove challenging, a fiber-based design could alleviate the $n^2$ wiring/packaging problem of many existing tactile glove efforts. To achieve the short focal distance, a \sensor{} skin design may take advantage of a microlens approach.

As a final note, we remark that the prototype presented here achieves miniaturization by
virtue of moving the illumination and imaging components outside the fingertip. A consequence of this decision is that the fingertip now has space that can be reclaimed. One possible use is to add additional sensors (e.g. temperature, vibration), thereby allowing the fingertip to approximate the multimodal nature of human fingertip sensing.


\section{Conclusion}
\label{sec:conclusion}

	As the world increasingly turns to automation to meet its needs, it is more vital than ever to empower robots with the \textit{senses} and thereby \textit{skills} to truly be helpful. Svelte tactile sensors will prove important for achieving human-level manipulation skill for robots, especially for tasks in constrained spaces. Towards this goal, we present a new framework for shrinking vision-based tactile sensors through the use of optical fiber bundles. We provide a proof of concept of this design philosophy with \sensor{}, a compact fingertip-shaped tactile sensor that is less than the width of a fifth percentile adult male finger~\cite{garrett1971adult}.

In this work, fundamental design considerations have been investigated for an fiber-based tactile sensor, including fiber illumination and fiber spatial resolution. The sensor image-to-force regression model shows good accuracy for a given indentation, illustrating a potential for use in closed-loop manipulation tasks in constrained environments. The image-to-hardness classification models demonstrate the potential for a miniature high-resolution tactile sensor in tissue palpation through internal orifices for patient examinations, such as for the cervix or the prostate. We open-source the design and manufacturing process of \sensor{} at \rev{\url{https://github.com/facebookresearch/digit-design}, and provide a video summary at \url{https://youtu.be/JaLls5nyrX4}}. 

While the medical applications for such a sensor are promising, we also want to acknowledge another blue-sky motivation for a slim, sensitive fingertip --- to enable a more human-like fluidity and grace from robots. If robots are ever to truly enter our daily lives and be useful with their hands, they must achieve both function \textit{and} form. In pursuit of the day when unnaturally blocky and bulky robot fingertips are no longer the standard (\fig{fig:demo}), optical fiber conduits are a promising path that we hope sparks future innovation.


\section*{Acknowledgement}
\begin{figure*}[t]
\centering
\includegraphics[width=0.85\textwidth]{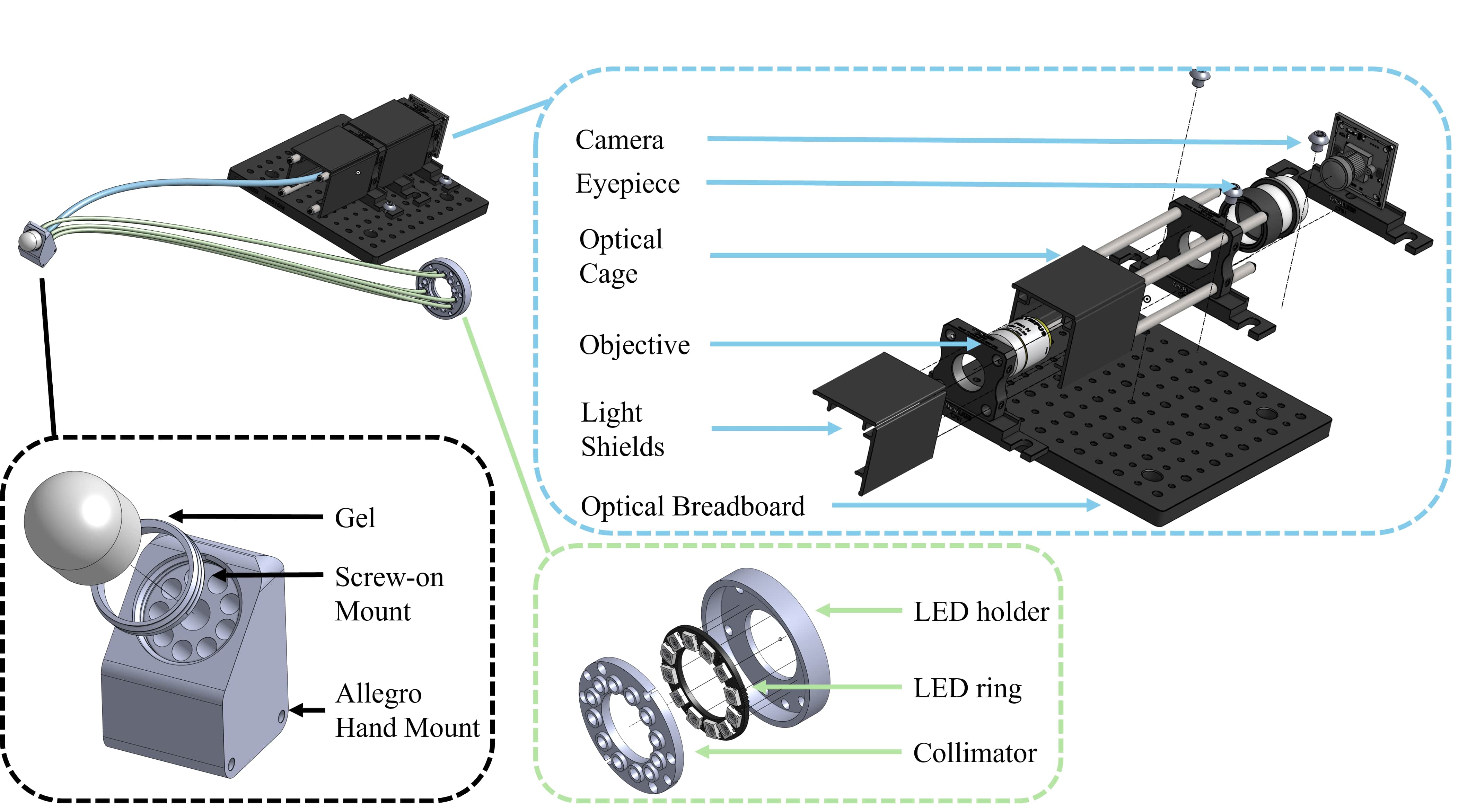}
\caption{CAD model of the \sensor{} proof of concept used in this work showing exploded views of the sensing element, imaging system, and illumination system.}
\label{fig:CAD}
\end{figure*}

We thank Raymond Santos for handling administrative orders and encouragement for J.D. while at Meta, and Thomas Craven-Bartle for helpful discussions on optics theory. 
We thank Lee White, MD, PhD and Lori Guelman, MSN, RN, FNP-BC for helpful discussions on clinical urology practice and feedback on the prostate phantoms, and Eugene Shkolyar, MD and Geoffrey Sonn, MD for allowing J.D. to shadow several radical prostatectomy cases. 
We also thank Jun En Low for providing a spare Android phone for the portable \sensor{} setup.
The design, fabrication, and data collection of the sensor was supported through a Meta internship. 
This work was partly supported by the German Research Foundation (DFG, Deutsche Forschungsgemeinschaft) as part of Germany’s Excellence Strategy – EXC 2050/1 – Project ID 390696704 – Cluster of Excellence “Centre for Tactile Internet with Human-in-the-Loop” (CeTI) of Technische Universität Dresden, and by Bundesministerium für Bildung und Forschung (BMBF) and German Academic Exchange Service (DAAD) in project 57616814 (\href{https://secai.org/}{SECAI}, \href{https://secai.org/}{School of Embedded and Composite AI}). 
We thank the Zentrum für Informationsdienste und Hochleistungsrechnen (ZIH) at TU Dresden for providing computing resources.


{\appendices
\section{Complete CAD Assembly}
\label{app:CAD}

The full CAD model for the \sensor{} setup is provided in \fig{fig:CAD}. \rev{The CAD files are also open-sourced at \url{https://github.com/facebookresearch/digit-design}.}

\section{\rev{Assembly Considerations and Modifications}}
\label{app:assembly}

\rev{As previously mentioned, the screw-on gel was made possibly because of the modularity of this fiber-based design, which allowed the circuitry to be entirely separated from the gel tip. This design was especially useful in a medical context because the gel tip could be easily replaced after each use. There was a concern that because the gel was not directly overmolded to the surface, the connection to the rest of the sensor system was not as secure. Depending on the tolerance of the 3D-printed parts, some cheaper printers were not able to print a fine enough thread, no such issues were noticed the final prototype. The following settings were used for all 3D-printed parts:}

\begin{table}[h]
    \rowcolors{2}{}{lightgray}
    \centering
    \caption{3D printer settings for printed parts in final prototype}
    \begin{tabular}{c|c}
         \hline
         Component & Specification \\
         \hline
         3D Printer & Stratysys J750 \\
         Layer Height & Default \\
         Material & Vero Black Plus \\
         Finish & Glossy \\
         \hline
    \end{tabular}
\end{table}

\rev{In addition, once the prototype was assembled, the image focus could be adjusted in a number of ways. The mounts for the microscope eyepiece and lens are mounted directly to the optical breadboard with metric screws; however, a 3D-printed part was designed with slots instead of mounting holes such that the pieces could slightly slide forward or backward to account for any misalignment before being tightened to a final position on the optical breadboard base. This allowed the exact placement of each component in the imaging assembly to be manually adjusted, and these 3D-printed mount designs are open-sourced. The adjustable diopter could also be used to adjust focus, but only by a small amount because of the shorter travel, so we relied on lens placement for any greater adjustments. A ThorLabs aluminum alignment target was used to help with adjustments, but because the process was manual, slight variations in focus occurred.}

\rev{To bring \sensor{} to the Stanford Hospital for the prostatectomy observation, it was required that all instrumentation remain relatively sterile/clean-able and strictly handheld. The design was made portable by placing the imaging assembly and lighting in a small plastic bin that could carried in and covered with a sterile bag. A hole was punched through the container to feed the fibers through, a second hole was punched through to feed the USB cable. All potential blood-contacting surfaces were covered with nitrile glove material. Because of the tight quarters of the operating room, image data was collected using a USB camera phone app on an Android device rather than a computer, but the image resolution and sampling frequency rate was matched to that of the benchtop data collection setup.}

\section{Sensor Construction Manufacturing Guide}
\label{app:manufacturingGuide}
A manufacturing guide and full bill of materials is also provided at \rev{\url{https://github.com/facebookresearch/digit-design}}.

\section{\rev{Robot Indenter Data Collection Setup}}
\label{app:indentersetup}
\rev{
To collect force data, several other vision-based tactile sensors used a ground truth sensor embedded under the tactile sensor prototype, so that the forces from the indenter probe onto the tactile sensor were measured. As a quasi-static system, a rigid frame transformation could be calculated between the tactile sensor and the ground truth sensor. 
}

\rev{In this work, the ground truth sensor could have been placed under the \sensor{} on the robot arm end effector. However the extruding fibers of \sensor{} made it not so straightforward to mount a ground truth sensor underneath, leading to an alternative configuration that instead placed the ground truth sensor under the indenter probe. In this case, the ground truth sensor measured the forces imparted on the indenter probe by the \sensor{}, which was translated into the probe using the Meca500 robot arm. The force balance equation was still $F_{\text{probe}} + F_{\text{pinki}} + F_{\text{endeffector}} + F_{\text{gravity}} = 0$. Because the axes of \sensor{} and the ground truth sensor were aligned, and only translation was applied, the frame transformation is straightforward to calculate.}

\section{Silicone Preparation for Hardness Experiments}
\label{app:hardnessPrep}

\begin{figure*}[t]
\centering
\subfloat[]{\includegraphics[width=0.33\textwidth]{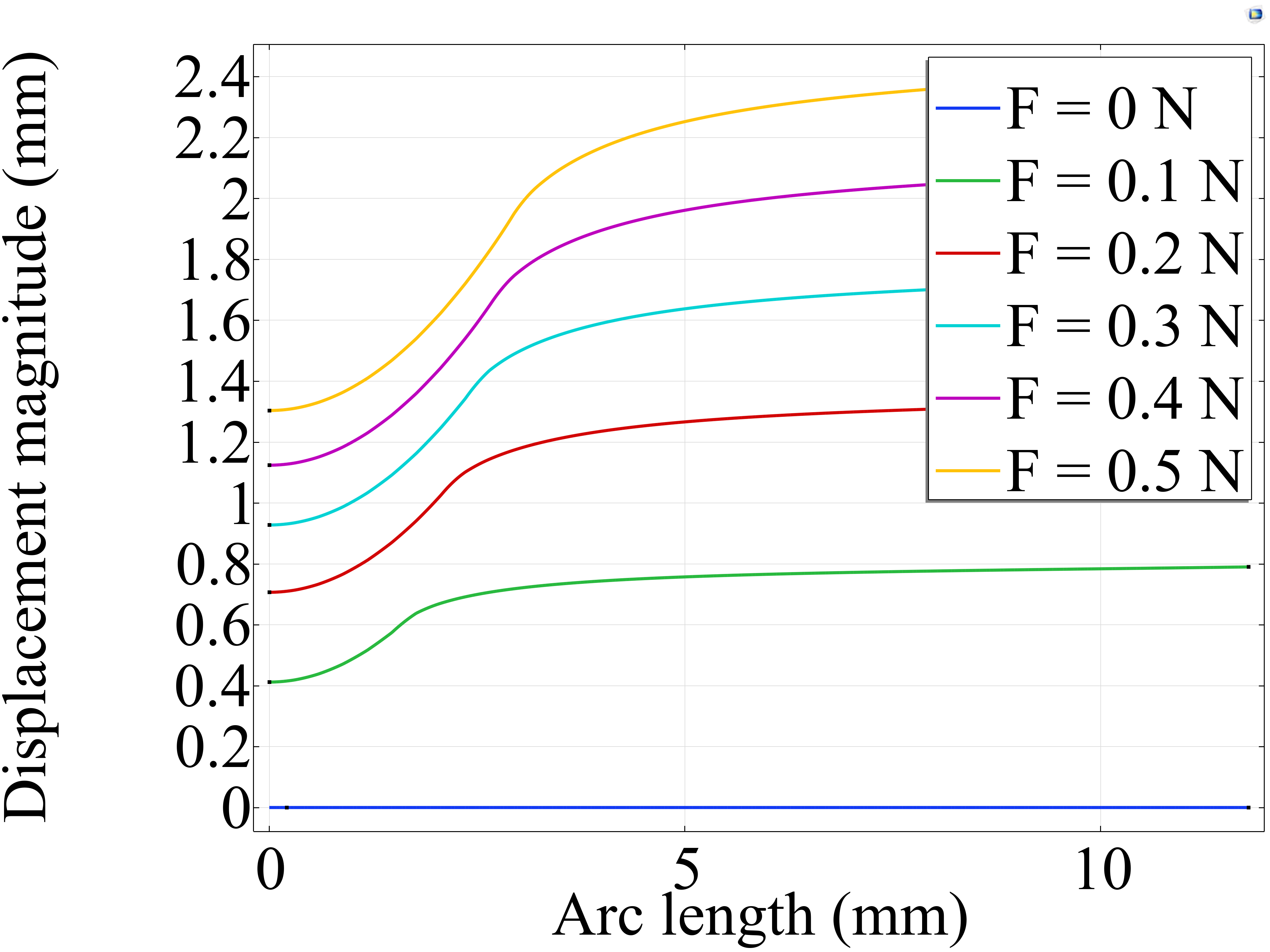}%
\label{fig:ecoflex-ecoflex-displacement}}
\hfil
\subfloat[]{\includegraphics[width=0.33\textwidth]{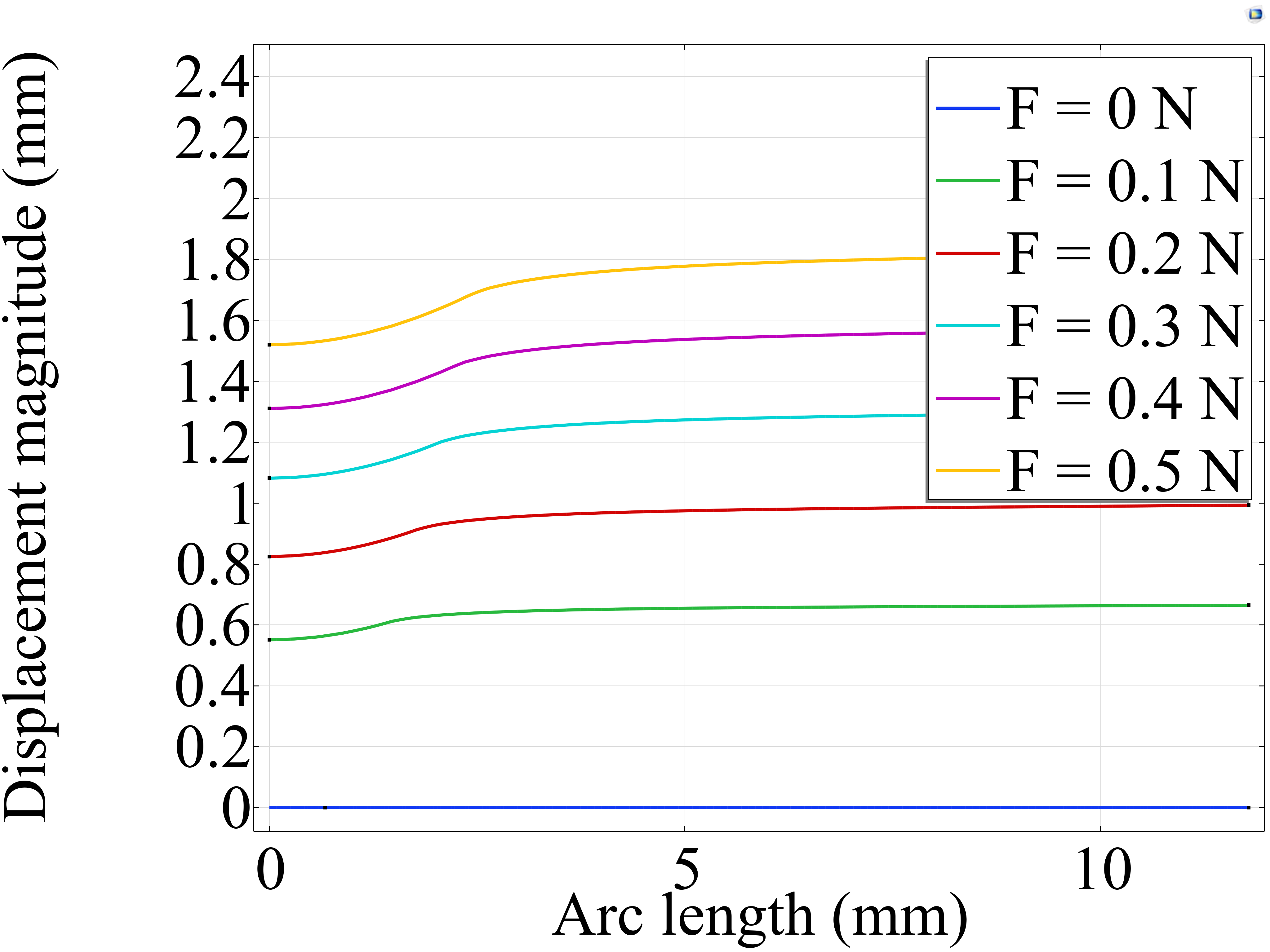}%
\label{fig:solaris-ecoflex-displacement}}
\hfil
\subfloat[]{\includegraphics[width=0.33\textwidth]{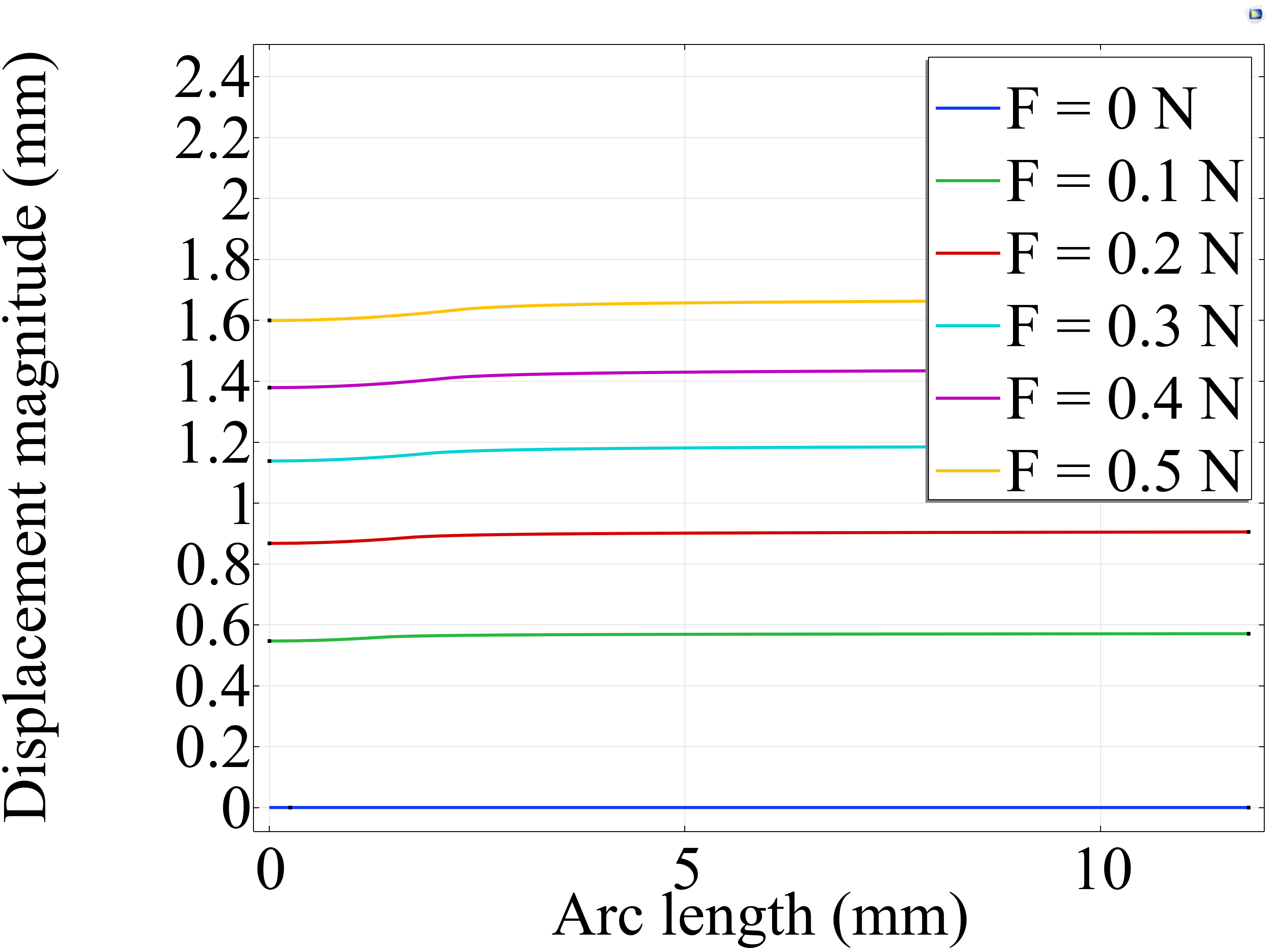}%
\label{fig:sortaclear40-ecoflex-displacement}}
\hfil
	\caption{\secondrev{Sensor surface displacement and therefore sensor sensitivity decreases with greater hardness mismatch. Examined are three scenarios: (\subref{fig:ecoflex-ecoflex-displacement}) no hardness difference between indenter and sample, (\subref{fig:solaris-ecoflex-displacement}) 10 degree Shore 00 hardness difference (current sensor), (\subref{fig:sortaclear40-ecoflex-displacement}) 36 degree Shore 00 hardness difference. }}
\label{fig:displacementvsforcepermaterial}	
\end{figure*}

\begin{figure*}[t]
\centering
\subfloat[Accuracy scores on silicone hardness data]{\includegraphics[width=0.33\textwidth]{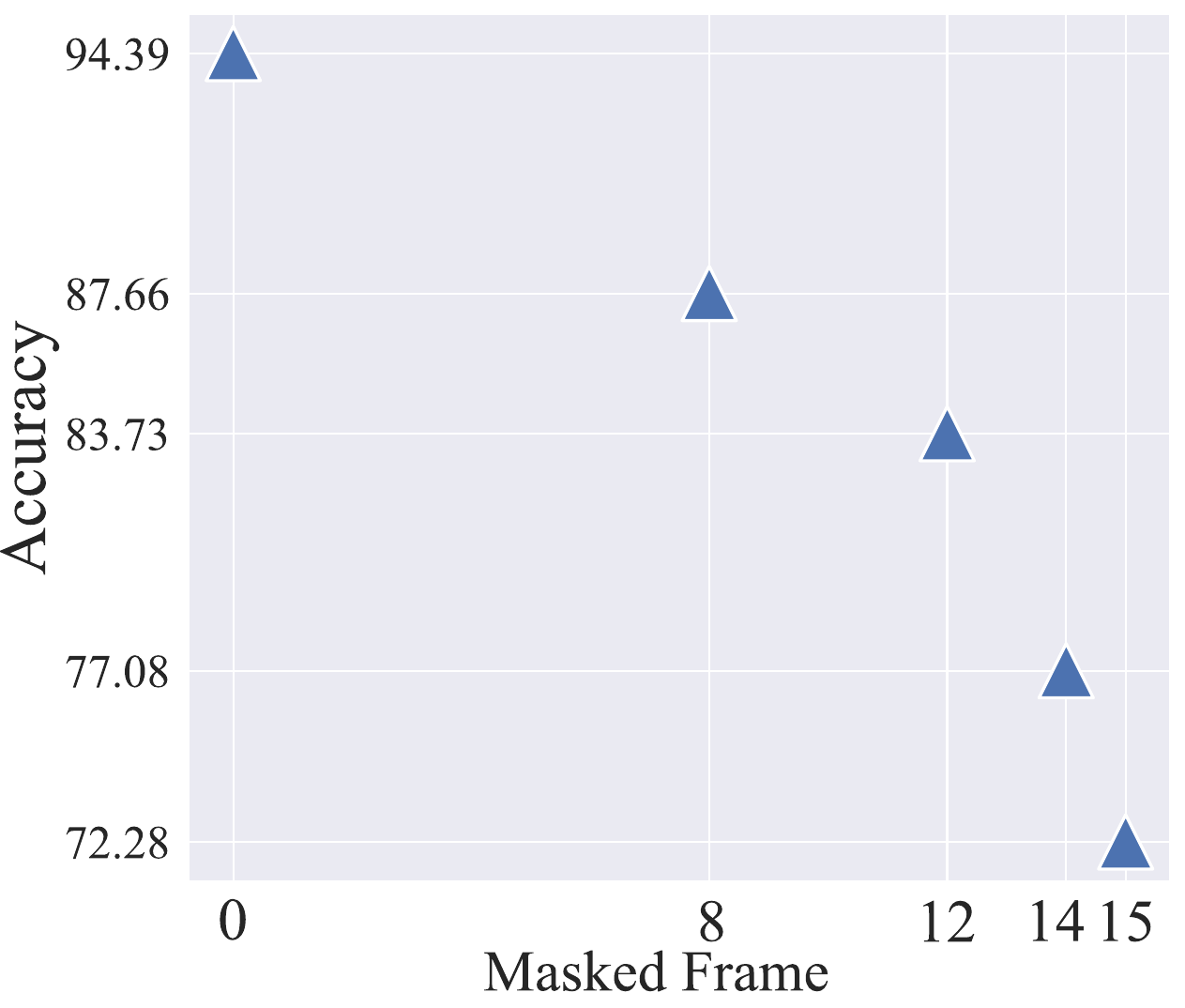}%
\label{fig:siliconeMaskAccuracy}}
\hfil
\subfloat[F1 scores on phantom hardness data]{\includegraphics[width=0.33\textwidth]{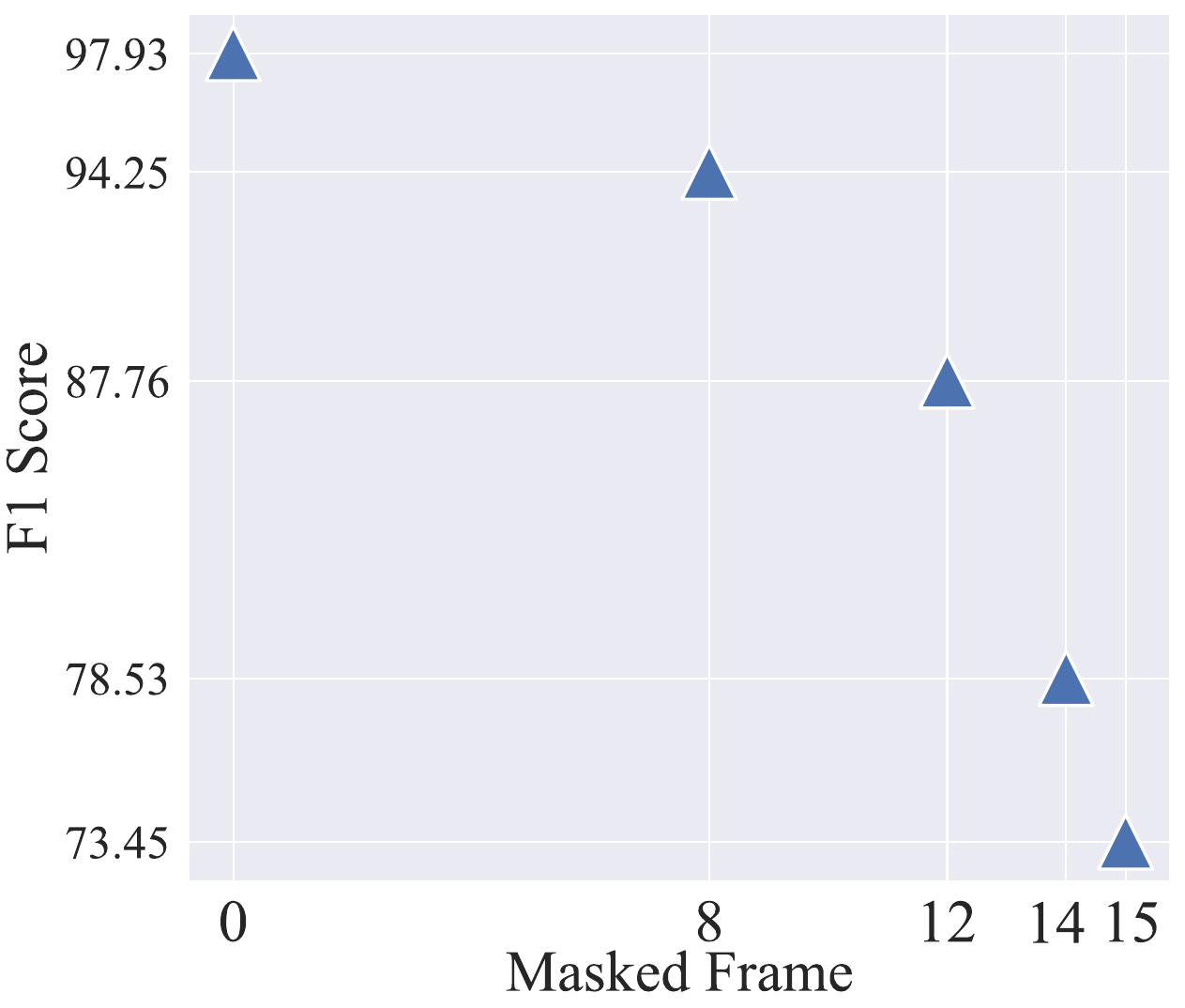}%
\label{fig:phantomMaskF1}}
\hfil
\subfloat[Accuracy scores on phantom hardness data]{\includegraphics[width=0.33\textwidth]{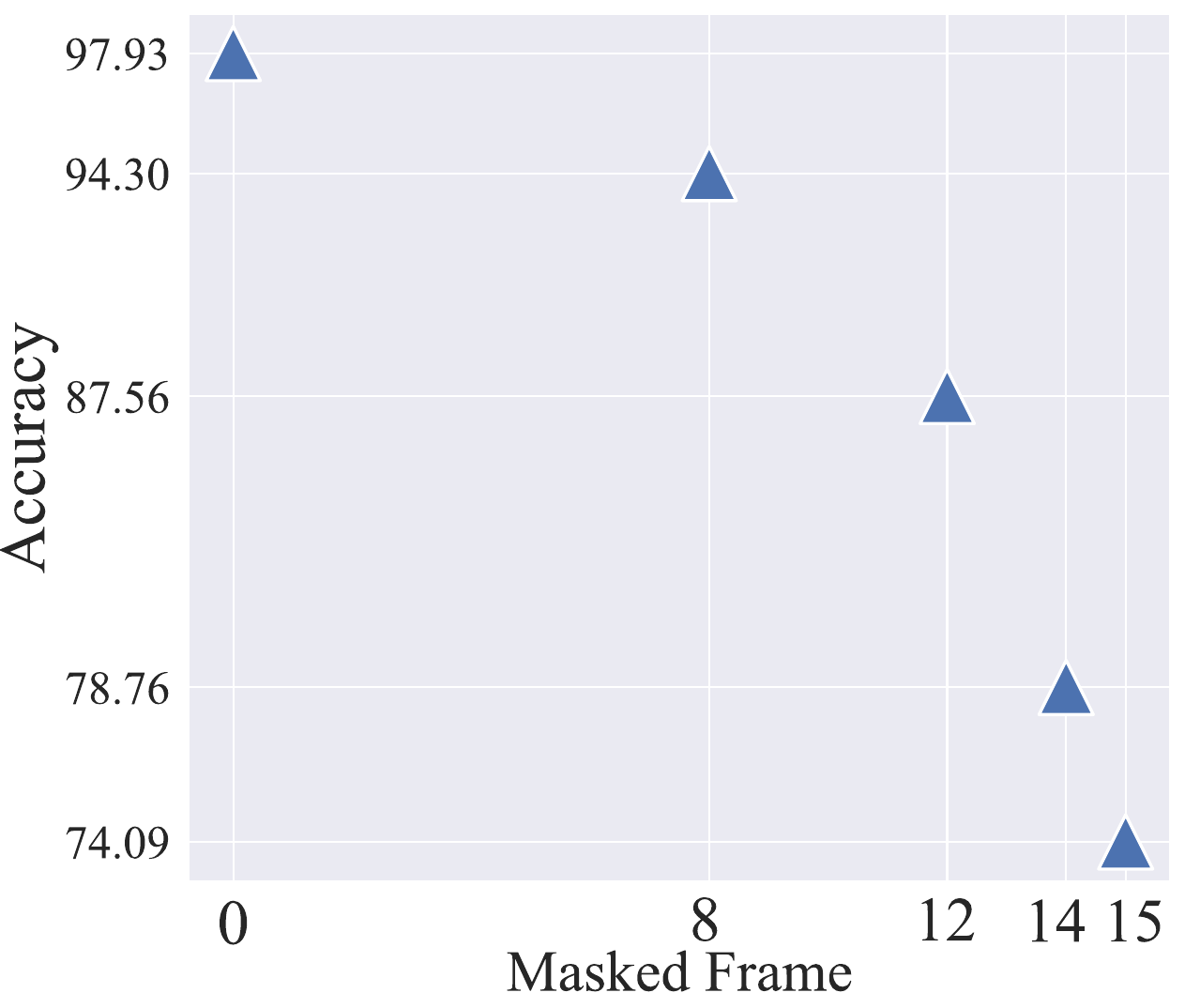}%
\label{fig:phantomMaskAccuracy}}
\hfil
	\caption{\rev{(\subref{fig:siliconeMaskAccuracy}) Accuracy scores for silicone sample hardness prediction task with delta-frames as input. Using all 16 frames, with no masked frames, leads to the highest F1 score. F1 scores decrease as more delta-frames are masked in the sequence, showing that longer sequences lead to better performance.(\subref{fig:phantomMaskF1}) The F1 scores for the image sequence length study on the phantom hardness prediction task with delta-frames as input. (\subref{fig:phantomMaskAccuracy}) Accuracy scores on phantom hardness prediction task with delta-frames as input.}}
\label{fig:appendixMaskResults}	
\end{figure*}

We include in \cref{tab:hardnessRatios} the mixing ratios used to achieve the silicone samples of different hardness values for this paper. After selecting the desired range of hardness values, we specifically chose to use Smooth-Sil 945 and Ecoflex 0050 for manufacturing ease: they both have similar cure times and one-to-one mixing ratios between Parts A and B. 

To determine the hardness value of mixed silicones, we purchased a commercial, portable Shore A durometer (Model: WonVon QMLBH0730HA-C-D) and took the average of five measurements on the flat silicone sample as the hardness value for that particular batch of samples. We report one sample as 50 on the Shore 00 scale as based on the manufacturer datasheet (the handheld Shore A scale durometer read 0.5 in manual measurements). 

\begin{table}[h]
\rowcolors{4}{}{lightgray}
\centering
    \caption{Mixing Ratios for Different Silicone Hardness}
    \label{tab:hardnessRatios}
\begin{tabularx}{\columnwidth}{|X|X|X|}
  \hline
  Hardness Value  & Smooth-On Ecoflex 0050 (\SI{}{\gram})  & Smooth-On Smooth-Sil 945 (\SI{}{\gram})\\
  \hline
  50 (Shore 00) & 16 & 0 \\
  1.5 (Shore A) & 14 & 2 \\
  5 (Shore A) & 12 & 4 \\
  6.5 (Shore A) & 11.5 & 4.5 \\
  11 (Shore A) & 10 & 6 \\
  16 (Shore A) & 8 & 8 \\
  20 (Shore A) & 7 & 9 \\
  25 (Shore A) & 6 & 10 \\
  30 (Shore A) & 5 & 11 \\
  35 (Shore A) & 4 & 12 \\
  42 (Shore A) & 2 & 14 \\
  45 (Shore A) & 0 & 16 \\
  \hline
\end{tabularx}
\end{table}

\section{\secondrev{Spatial Points For Study of Separated Locations}}
\label{app:differentspatiallocations}

\secondrev{We provide here the separated train/val/test locations for the smaller 10k dataset study. Within the region centered on the sensor tip, the validation spatial points for the \SI{4}{\milli\meter}, \SI{12}{\milli\meter}, and Flat indenters were $(-1,0)$, $(-3.3,0.2)$, and $(1.2,-1.3)$ respectively; for the test set points, they were $(-1, 1)$, $(-2.3,0.2)$, and $(-1,0)$ respectively. The training points were a sweep along the x-axis at increments of \SI{0.1}{\milli\meter} from $(-0.9,0)$ to $(1,0)$.}

\section{\rev{COMSOL FEA Simulation Parameters}}
\label{app:FEA}

\rev{Figure~\ref{fig:FEA} includes the results of the FEA plots which studied an adversarial case when the contact patch size is the same. The modeling was done in COMSOL Multiphysics 6.2. The exact geometries of sensor tip and the silicone samples were used. Because the scenario was radially symmetric, only one half of the indenter and sample was modeled to save computation time. The indenting tip was modeled with a Young's modulus of \SI{360}{\kilo\pascal}, a Poisson's ratio of 0.49, and a density of \SI{1001}{\kilo\gram/\meter^3} based on previously published values for Solaris \cite{darby2022modulus}. \secondrev{The Ecoflex 00-50 sample was modeled with a Young's modulus of \SI{0.09}{\mega\pascal}, a Poisson's ratio of 0.36, and a density of \SI{1.07}{\gram/\centi\meter^3}~\cite{lavazza2020study}.} When indenting, the sensor tip applied a ramped load up to \SI{0.5}{\newton} normally into the sample. The indenter tip contact boundary is meshed at 0.3 and the sample contact boundary is meshed at 0.15.}

\section{\secondrev{Additional FEA Study on Tradeoff of Mismatched Indenter and Sample Hardnesses}}
\label{app:stiffnessminimumforce}

\secondrev{As noted in \cref{subsec:ablation}, the \sensor{} material was not optimized for the softest tested samples, as there was an on the order of 10 degree difference in Shore  hardness between the indenter (Shore 10A or about Shore 00-60) and the softest sample (Shore 00-50). This difference was not significant enough to degrade the performance on the softest samples in this work. However, previous research on tactile hardness estimation  has shown that when there was a more significant mismatch between the indenter and sample hardness, the accuracy of the hardness estimation was reduced~\cite{yuan2017shape}. Previous research has also shown that stiffer gel construction can be more robust against abrasion~\cite{lambeta2020digit}, attractive for a critical application such as medicine. To better indicate the price of using a stiffer-than-optimal indenter material on soft samples, we report an additional FEA study. The study compared the deformation at the surface for presses of 0 to \SI{0.5}{\newton} under three scenarios: (a) an Ecoflex 00-50 indenter, (b) the \sensor{} (Shore 10A or Shore 00-60) indenter, (c) and a plausible SORTA-Clear 40 (Shore 40A or Shore 00-86) indenter all on a \SI{12}{\milli\meter} diameter Ecoflex 00-50 sample. Based on the results, shown in \cref{fig:displacementvsforcepermaterial}, the (a) impedance-matched scenario is the most sensitive; a light indentation of \SI{0.1}{\newton} causes a surface displacement of about \SI{0.22}{\milli\meter}. To achieve the same displacement, the (b) current sensor setup must be pressed with \SI{0.5}{\newton}, showing that the mismatch in material does lead to a decrease in sensitivity. In scenario (c) where the mismatch approached that of a 50x magnitude difference, there was practically no deformation for light forces. This result suggested that increasing palpation force could help compensate for small mismatches in material hardness on the order of 10 degrees, but not for larger mismatches.}

\section{\rev{Additional Results From Studying Image Sequence Length}}
\label{app:ablation}

\rev{Figure~\ref{fig:appendixMaskResults} show the results for the delta-frame study on the importance of image-sequence length to hardness prediction success, where the delta-frames were computed as $\Delta_{l+1,l} = (\text{frame}_{l+1} - \text{frame}_l)$. This study reported the accuracy scores on silicone hardness data, and both the F1 score and accuracy scores for the phantom hardness data. The results indicated that longer image sequences were important for the model to accurately predict hardness.}

\section{\rev{Hyperfisheye Lens Design Considerations for Miniature Tactile Sensors}}
\label{app:hyperfisheye}

\begin{figure}[]
    \centering
    \includegraphics[width=0.7\columnwidth]{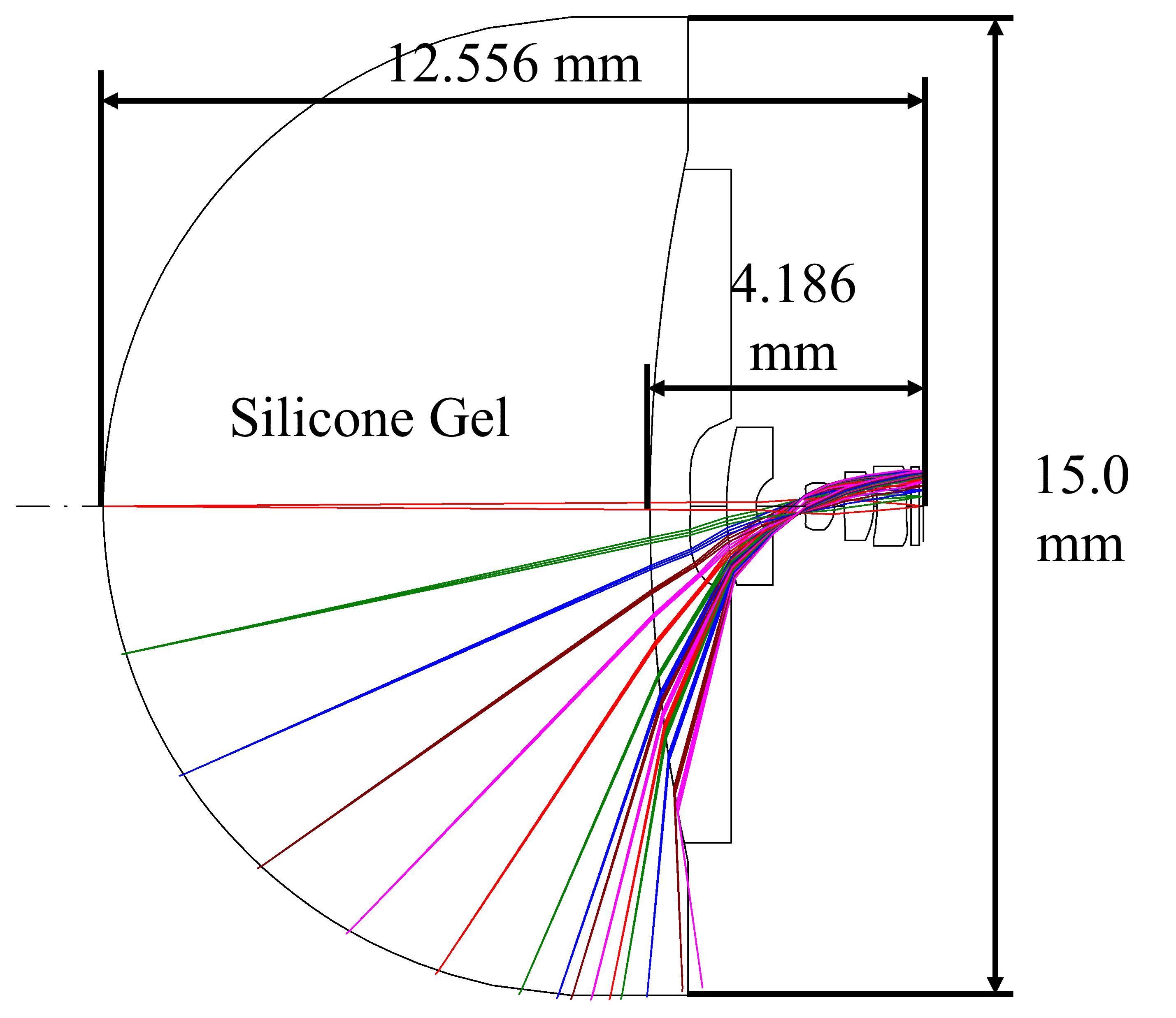}%
    \caption{A theoretical custom distal hyperfisheye lens design for a \SI{15}{\milli\meter} diameter dome-shaped silicone gel. The lens design has 5 components and provides a \SI{192}{\degree} field of view.}
    \label{fig:lensDesign}
\end{figure}

\begin{figure}[]
    \centering
    \includegraphics[width=1\linewidth]{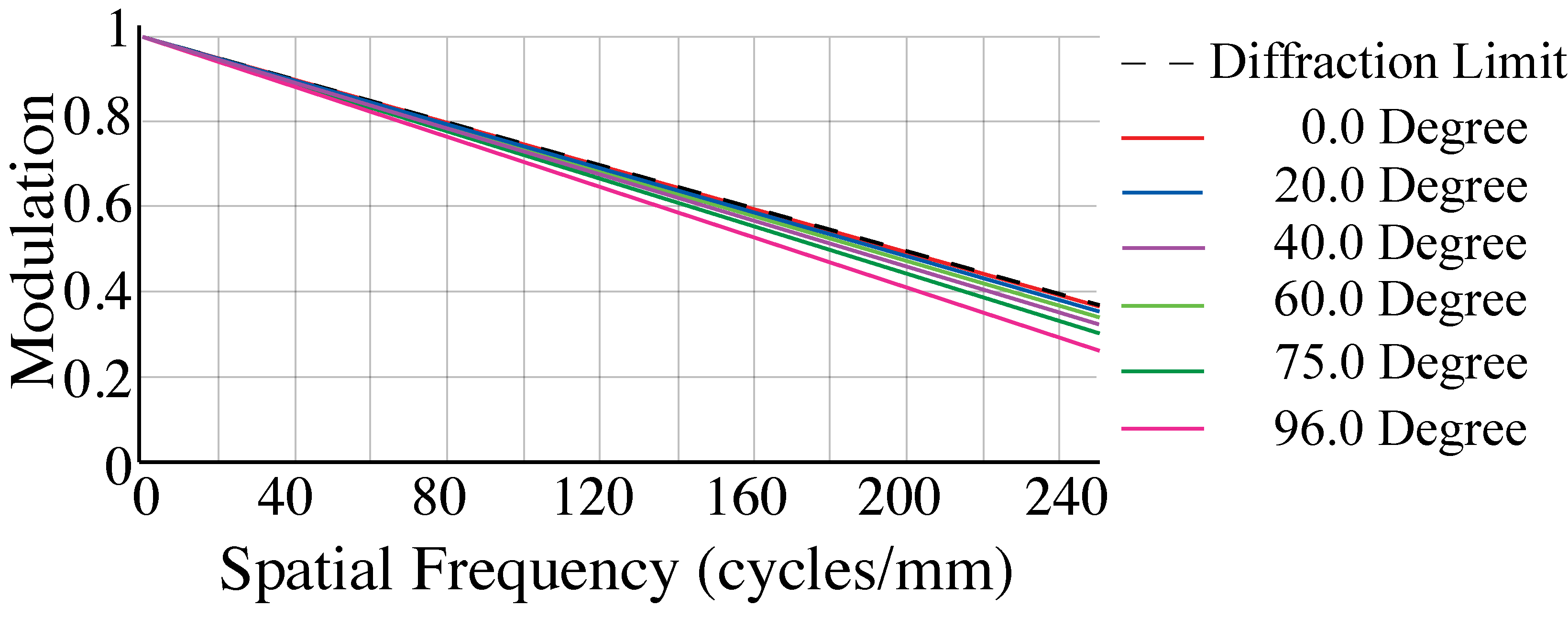}%
    \caption{\rev{The Modulation Transfer Function (MTF) curves for the custom distal hyperfisheye lens design show that there is only a 15\% decrease in performance for indentations at the edges of the sensor compared to those at the tip.}
    }
    \label{fig:MTF}
\end{figure}

\rev{There are some considerations that must be made to design a compact hyperfisheye lens specifically for tactile sensing, such as the theoretical \SI{192}{\degree} field-of-view lens shown in \fig{fig:lensDesign}. For a dome-shaped sensor made of bulk silicone, the lens must be modeled for a different index of refraction than air. Compared to commercially-available lenses intended for photography, lenses for tactile sensing are relatively unconcerned about optical distortions. Furthermore, intentionally allowing chromatic aberrations and removing anti-reflective coatings better captures reflection and scattering inside the elastomer. The design modeled here uses APEL 5014 plastic material and could be manufactured either by diamond turning or molding.}

\rev{The MTF performance curves are modeled in Code V and illustrated in \cref{fig:MTF}, relating contrast over spatial frequencies (resolution) at the \SI{587.6}{\nano\meter} wavelength. As a lens for a dome-shaped tactile sensor, the design optimizes for high MTF across the field of view. As seen in the MTF curves, the imaging resolution for an indentation at the sides will be slightly worse than the performance for an indentation at the top. For example, the worst resolving performance (about 170 cycles/\SI{}{\milli\meter}) occurs at the maximal diffraction angle of \SI{96}{\degree}, corresponding to an indentation at the very edge of the field of view. However, when compared with the resolving performance for an indentation at the top of the sensor (about 200 cycles/\SI{}{\milli\meter}), this is only about a 15\% reduction in resolution.}

\rev{Due to resource constraints, we were unable to manufacture the hyperfisheye lens design in time for testing. Although the custom lens design presented  was not integrated in the sensor system, the MTF analysis presented should accurately reflect the real-world MTF when manufacturing tolerances are performed correctly.
While the spatial resolution result reported in \cref{sec:characterization} is with the \SI{60}{\degree} lens, the same resolution with a hyperfisheye lens could be attained by increasing the fiber core count accordingly.}



\bibliographystyle{IEEEtran}
\bibliography{paper}


\section*{Biography Section}
\vspace{-33pt}
\begin{IEEEbiography}[{\includegraphics[width=1in,height=1in,clip,keepaspectratio]{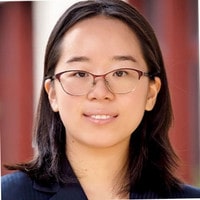}}]{Julia Di}
received the B.S. degree in Electrical Engineering with a minor in Computer Science from Columbia University, New York, NY, USA in 2018. She received the M.S. in Mechanical Engineering from Stanford University, Stanford, CA, USA in 2020. Her research interests include tactile sensing, sensor networks, and perception of grasping. 
\end{IEEEbiography}

\vspace{-55pt}
\begin{IEEEbiography}[{\includegraphics[width=1in,height=1in,clip,keepaspectratio]{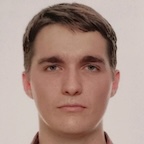}}]{Zdravko Dugonjic}
received the MSc degree in Computer Science from Université Grenoble Alpes in 2023.  He is a Ph.D. student in Learning, Adaptive Systems, and Robotics (LASR) Lab, at Technische Universit\"at Dresden. His research interests are touch perception and machine learning.
\end{IEEEbiography}

\vspace{-55pt}
\begin{IEEEbiography}[{\includegraphics[width=1in,height=1in,clip,keepaspectratio]{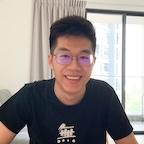}}]{Will Fu}
received a Bachelor Degree in Computer Science and a Bachelor Degree in Business from Nanyang Technological University in 2017. His research interests are computer vision for robotics and deep learning recommendation models.
\end{IEEEbiography}

\vspace{-55pt}
\begin{IEEEbiography}[{\includegraphics[width=1in,height=1in,clip,keepaspectratio]{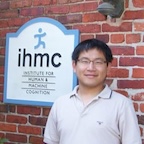}}]{Tingfan Wu}
received the Ph.D. degree in Computer Science from University of California, San Diego, San Diego, CA, USA in 2013. Dr. Wu is currently a Research Engineer at Meta, Menlo Park, CA, USA. His research interests include machine learning and robotics.
\end{IEEEbiography}

\vspace{-55pt}
\begin{IEEEbiography}[{\includegraphics[width=1in,height=1in,clip,keepaspectratio]{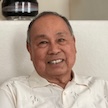}}]{Romeo Mercado}
received the Ph.D. degree in Optical Science from University of Arizona in 1973. Dr. Mercado is an optical scientist and consultant with over 30 years of experience in optics. He holds 35 U.S. patents and a number of foreign patents on optical system designs.
\end{IEEEbiography}

\vspace{-55pt}
\begin{IEEEbiography}[{\includegraphics[width=1in,height=1in,clip,keepaspectratio]{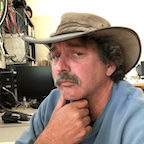}}]{Kevin Sawyer} received the Ph.D. degree in Engineering Mechanics from the University of Arizona in 1995. Dr. Sawyer is a senior optomechanical engineer and adjunct professor at San Jose State University, with over 30 years of experience in optoelectronics.
\end{IEEEbiography}

\vspace{-55pt}
\begin{IEEEbiography}[{\includegraphics[width=1in,height=1in,clip,keepaspectratio]{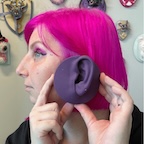}}]{Victoria Rose Most}
received the B.A. in Animation from the College for Creative Studies in 2007. She is currently a Design Model Maker at Meta, Menlo Park, CA, USA. She has over 15 years of experience in silicone molding and casting in the entertainment industry, including Coraline (2009). 
\end{IEEEbiography}
\vspace{11pt}

\vspace{-55pt}
\begin{IEEEbiography}[{\includegraphics[width=1in,height=1in,clip,keepaspectratio]{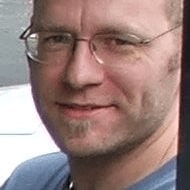}}]{Gregg Kammerer} received the B.S. degree in Law Enforcement and Justice Administration from Western Illinois University in 1994. He has 30 years of experience as a model maker and machinist prototyper.
\end{IEEEbiography}
\vspace{11pt}

\vspace{-55pt}
\begin{IEEEbiography}[{\includegraphics[width=1in,height=1in,clip,keepaspectratio]{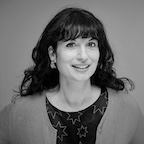}}]{Stefanie Speidel} (Senior Member, IEEE) received the Ph.D. (Dr.Ing.) degree from the Karlsruhe Institute of Technology (KIT), Karlsruhe, Germany, in 2009. She has been a Professor for “Translational Surgical Oncology” at the National Center for Tumor Diseases (NCT/UCC) Dresden, Dresden, Germany, since 2017, and a Deputy Speaker of the DFG Cluster of Excellence “Centre for Tactile Internet with Human-in-the-Loop” (CeTI) since 2019 and the Konrad Zuse AI school SECAI since 2022.
\end{IEEEbiography}

\vspace{-55pt}
\begin{IEEEbiography}[{\includegraphics[width=1in,height=1in,clip,keepaspectratio]{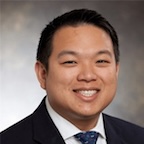}}]{Richard E. Fan} received the Ph.D. degree in biomedical engineering from University of California, Los Angeles in 2010. Dr. Fan is currently the Engineering Director of the Urologic Cancer Innovation Lab and a Clinical Assistant Professor in Urology at Stanford University.
\end{IEEEbiography}

\vspace{-55pt}
\begin{IEEEbiography}[{\includegraphics[width=1in,height=1in,clip,keepaspectratio]{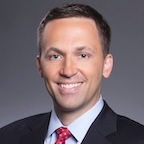}}]{Geoffrey Sonn}, MD is a board certified urologist who specializes in treating patients with prostate and kidney cancer. Dr. Sonn has a particular interest in cancer imaging, MRI-Ultrasound fusion targeted prostate biopsy, prostate cancer focal therapy, and robotic surgery for prostate and kidney cancer. He is a member of Stanford Bio-X, the Stanford Cancer Institute, and Associate Member of the Canary Center at Stanford for Cancer Early Detection.
\end{IEEEbiography}

\vspace{-55pt}
\begin{IEEEbiography}[{\includegraphics[width=1in,height=1in,clip,keepaspectratio]{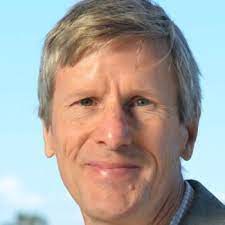}}]{Mark R. Cutkosky} received the Ph.D. degree in mechanical engineering from Carnegie Mellon University, Pittsburgh, PA, USA in 1985.\\
He is the Fletcher Jones Professor in Mechanical Engineering at Stanford University, Stanford, CA, USA. His research interests include bioinspired robots, haptics, and rapid prototyping processes.\\
Dr. Cutkosky is a Fellow of IEEE and ASME, and an IEEE RA-L Pioneer in robotics.
\end{IEEEbiography}

\vspace{-55pt}
\begin{IEEEbiography}[{\includegraphics[width=1in,height=1in,clip,keepaspectratio]{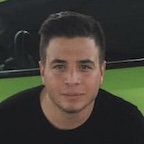}}]{Mike Lambeta} is an AI Research Engineer in Hardware at Meta, Menlo Park, CA, USA. His research interests include tactile sensing and robotics.
\end{IEEEbiography}

\vspace{-55pt}
\begin{IEEEbiography}[{\includegraphics[width=1in,height=1in,clip,keepaspectratio]{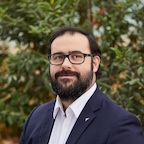}}]{Roberto Calandra}
is a Full (W3) Professor at the Technische Universit\"at Dresden. Previously, he founded at Meta AI (formerly Facebook AI Research) the Robotic Lab in Menlo Park. Prior to that, he was a Postdoctoral Scholar at the University of California, Berkeley (US). His education includes a Ph.D. from TU Darmstadt (Germany), a M.Sc. in Machine Learning and Data Mining from the Aalto university (Finland), and a B.Sc. in Computer Science from the Università degli studi di Palermo (Italy). His scientific interests are broadly at the conjunction of Robotics, Touch Sensing, and Machine Learning. In 2024, he received the IEEE Early Academic Career Award in Robotics and Automation.
\end{IEEEbiography}



\end{document}